\documentclass[a4paper,10.5pt]{elsarticle}

\usepackage{lineno,hyperref}
\usepackage{amsmath}
\usepackage{amsfonts}
\usepackage{amssymb}
\usepackage{pdflscape}
\usepackage{threeparttable}
\usepackage{mathrsfs}
\usepackage{bm}
\usepackage{geometry}
\usepackage{rotating}

\usepackage[dvips]{epsfig}
\usepackage{epsfig, epsf}
\usepackage[dvips]{color}

\usepackage{algorithm}
\usepackage{algorithmic}
\usepackage{multirow}

\newcommand{\tabincell}[2]{\begin{tabular}{@{}#1@{}}#2\end{tabular}}

\usepackage{longtable}

\usepackage{booktabs}

\usepackage{subfig}

\modulolinenumbers[5]

\newdefinition{definition}{Definition}
\newtheorem{thm}{Proposition}
\newproof{pf}{Proof}

\usepackage{setspace}
\begin{document}

\begin{frontmatter}
	
	\title{Data-driven preference learning methods\\ for value-driven multiple criteria sorting with interacting criteria}
	
	\author[mymainaddress]{Jiapeng Liu\corref{mycorrespondingauthor}}
	\cortext[mycorrespondingauthor]{Corresponding author}
	\ead{jiapengliu@mail.xjtu.edu.cn}
	
	\author[mysecondaddress]{Mi{\l}osz Kadzi{\'n}ski}
	\ead{milosz.kadzinski@cs.put.poznan.pl}
	
	\author[mymainaddress]{Xiuwu Liao}
	\ead{liaoxiuwu@mail.xjtu.edu.cn}
	
	\author[mymainaddress]{Xiaoxin Mao}
	\ead{maoxiaoxin29@stu.xjtu.edu.cn}	
	
	\address[mymainaddress]{School of Management, Xi'an Jiaotong University, Xi'an, 710049, Shaanxi, P.R. China}
	
	\address[mysecondaddress]{Institute of Computing Science, Poznan University of Technology, Piotrowo 2, 60-965 Pozna{\'n}, Poland}
	
	\begin{abstract}
		The learning of predictive models for data-driven decision support has been a prevalent topic in many fields. However, construction of models that would capture interactions among input variables is a~challenging task. In this paper, we present a~new preference learning approach for multiple criteria sorting with potentially interacting criteria. It employs an additive piecewise-linear value function as the basic preference model, which is augmented with components for handling the interactions. To~construct such a~model from a given set of assignment examples concerning reference alternatives, we develop a convex quadratic programming model. Since its~complexity does not depend on the number of training samples, the proposed approach is capable for dealing with data-intensive tasks. To improve the generalization of the constructed model on new instances and to overcome the problem of over-fitting, we employ the regularization techniques. We also propose a few novel methods for classifying non-reference alternatives in order to enhance the applicability of our approach to different datasets. The practical usefulness of the proposed method is demonstrated on a~problem of parametric evaluation of research units, whereas its predictive performance is studied on several monotone learning datasets. The experimental results indicate that our approach compares favourably with the classical UTADIS method and the Choquet integral-based sorting model.
	\end{abstract}
	
	
\end{frontmatter}

\section{Introduction}

\noindent The purpose of multiple criteria sorting (also called ordinal classification) is to help a decision maker (DM) to assign a finite set of alternatives to pre-defined and preference ordered classes according to their performances on multiple criteria. In the past decade, sorting has been among the most growing areas in Multiple Criteria Decision Aiding (MCDA) for addressing problems in various disciplines such as credit rating, policy making and assessment, inventory control, project management, supplier segmentation, recommendation systems, risk assessment, or competitiveness analysis.

In the majority of recently proposed methods, dealing with sorting usually requires the DM to express his/her preferences in form of assignment examples concerning a subset of reference alternatives. Such information is used to construct a preference model compatible with the DM's preferences, which is subsequently employed for comparing the alternatives against some class profiles or for establishing the preference relation among alternatives in a way that allows to derive the class assignments.
In many such approaches, constructing a preference model is usually organized as a series of interactions in which the DM provides incremental preference information in order to calibrate the constructed model to better fit his/her preferences. Meanwhile, the DM could verify the consequences of the provided preference information on the decision outcomes, which allows him/her to shape one's preferences progressively and finally to be convinced about the validity of arrived sorting recommendation.

Nowadays, data-driven decision support has been an important issue for many businesses. Specifically, with the recent development of information technology, decision support systems are used to assist humans in deriving insights and making decisions through the analysis of increasingly complex data. For example, financial institutions develop the systems for evaluating credit risks of firms and individuals according to their transaction data and financial indicators and for deciding whether to grant a loan. Furthermore, firms rely on customer relationship management systems to construct profiles of customers from their on-line and off-line behaviours and perform market segmentation in order to tailor different marketing policies for targeted segments. Although these two real-world applications can be viewed in terms of sorting, an intrinsic distinction between such data-driven decision problems and traditional MCDA problems consists in the former requiring the preference discovery to be performed automatically without further intervention of the DM, whereas the latter expecting the DM's active participation in the preference construction process.

Preference discovery (also called preference learning) has been an important field in the Machine Learning (ML) community. Its primary focus is on constructing -- in an automatic way -- a model from a~given training sample and predicting preference for a new sample of data so that the outcomes obtained with the discovered model are ``similar'' to the provided training data in some sense. In contrast to MCDA, where the preference model is co-constructed with the participation of the DM so that to ensure its interpretability and descriptive character, preference learning in ML is mainly concerned about the ability of capturing complex interdependencies between the input and the output as well as the predictive performance of the discovered model. This difference is further reflected in the form of preference models employed in both fields. On the one hand, additive models are widely used in MCDA due to their advantage of intuitiveness and comprehensibility. On~the other hand, some non-linear models, such as kernel methods and neural networks, are often used in ML to capture interdependencies and other complex patterns in data. Although such non-linear models offer greater flexibility in terms of fitting the learning data and recognizing patterns, they are too complex to be interpreted by users, and therefore they are often referred to as ``black boxes''.

In this paper, we bridge the gap between the fields of MCDA and ML by proposing a new preference learning approach for data-driven multiple criteria sorting problems. We aim to learn a preference model from historical decision examples (also called training samples) so that it can be used to recommend a decision for the non-reference alternatives. The model should not only have a high predictive performance, but also allow for interpretable description of preferences. Specifically, the proposed approach can capture potential interactions among criteria, which is relevant for numerous real-world applications. For example, let us consider computers evaluated in terms of the number of CPU cores, CPU speed, and price. On the one hand, there may exist a~negative interaction between the number of CPU cores and CPU speed, because a computer that has a~large number of CPU usually has a high CPU speed. Thus, when considering such a pair of criteria jointly, its impact on the comprehensive quality of a computer should be lesser than a simple addition of the two impacts generated by considering each of the two criteria separately. On the other hand, there may exist a~positive interaction between CPU speed and price, because a high CPU speed usually comes along with a~high price. Thus, a computer with a high CPU speed and a low price is much appreciated, as the joint impact of such a~pair of criteria on the overall desirability of a computer should be larger than a simple summation of the two impacts viewed individually.

In MCDA, several models for capturing the interactions between criteria have been developed. Firstly, a~multi-linear utility function is a more general form of a~value function, which aggregates products of marginal utilities on each criterion over all subsets of criteria. Secondly, the Choquet integral can be seen as an average of marginal values according to a capacity that measures the relative importance of every subset of criteria. If there is a~positive (negative) interaction between two criteria, the weight assigned to such a pair is larger (smaller) than the sum of weights assigned to each of the two criteria separately. In particular has advocated the use of the Choquet integral as an~aggregation model for preference learning, and incorporated it within an extension of logistic regression. The main limitation of the two aforementioned models derives from the need of expressing the performances on different criteria on the same scale or bringing them to the joint scale by the use of marginal value functions which need to be specified beforehand. This poses a serious burden for the use of such preference models in real-world decision problems. The third type of a~preference model handling interactions between criteria is a general additive value function augmented by a pair components that capture the positive and negative interactions for pairs of criteria. The latter  model neither requires specification of all performances on the same scale nor a priori definition of marginal values. Its construction has been traditionally based on linear programming techniques used within an interactive procedure during which the DM could progressively discover the pairs of interacting criteria.

In the proposed preference learning approach, we consider an additive value model with piecewise-linear marginal functions under the preferential independence hypothesis, and then extend it for capturing the interactions. For this purpose, we adapt the preference model proposed in by means of two types of expressions for quantifying the positive and negative interactions among pairs of criteria. Consequently, our approach belongs to the family of value-based MCDA methods, which allow for establishing preference relations among alternatives by comparing their numerical scores, thus preserving the advantage of intuitiveness and comprehensibility. Our approach does not require all criteria to be expressed on the same scale, admitting an assessment of both the relative importances of criteria and the potential interaction intensities between pairs of criteria. 

We also introduce methodological advances in ML to enhance the predictive ability of the constructed preference model and the computational efficiency of the preference learning procedure. We formulate the learning problem in the regularization framework and use regularization terms for improving the generalization ability of the constructed model on new instances. Moreover, by utilizing the properties of value functions, we formulate a convex quadratic programming model for constructing the preference model. Since the complexity of this technique is irrelevant from the number of training samples, it is suitable for addressing data-intensive tasks and the respective models can be derived using popular solvers without extraordinary efforts. In addition, we propose four methods for classifying non-reference alternatives once the preference model with the optimal fitting performance is obtained. Consequently, the generalization performance can be improved by selecting one of the four procedures that proved to be the most advantageous for a given dataset.

The various variants of the proposed approach in terms of different interaction expressions and  methods for classifying non-reference alternatives are validated within an extensive computational study. In particular, the practical usefulness of the proposed method is demonstrated on a problem of parametric evaluation of research units. In this perspective, we discuss how to interpret information on the relative importance of criteria and the interaction coefficients between pairs of criteria. Moreover, we compare the proposed approach with the UTADIS method and the Choquet integral-based sorting model in terms of their predictive performances on nine monotone learning datasets. 

The remainder of the paper is organized in the following way. In Section \ref{sec-proposed-approach}, we present the learning approach for addressing sorting problems with potentially interacting criteria. In Section \ref{sec-experimental-analysis}, we apply the proposed approach to a~problem of parametric evaluation of Polish research units. We also discuss the experimental results derived from the comparison of the introduced method with UTADIS and the Choquet integral-based model on several public datasets. Section~\ref{sec-conclusions} concludes the paper and provides avenues for future research.

\section{Preference learning approach for sorting problems with potentially interacting criteria}
\label{sec-proposed-approach}

\subsection{Problem description}

\noindent We describe the considered sorting problems with the following notation:

\begin{itemize}
	\item ${A^R} = \left\{ {a,b,...} \right\}$ -- a~set of reference alternatives (training sample) for which the classification is known;
	\item $A = \left\{ {a_1,a_2,...} \right\}$ -- a~set of non-reference alternatives to be classified;
	\item $CL = \left\{ {C{l_1},...,C{l_q}} \right\}$ -- a~set of decision classes, such that $C{l_{s + 1}}$ is preferred to $C{l_s}$ (denoted by $C{l_{s + 1}} \succ C{l_s}$), $s = 1,...,q - 1$, and $Cl_1$ and $Cl_q$ are, respectively, the least and the most preferred ones;
	\item $G = \left\{ {{g_1}, ..., {g_n}} \right\}$ -- a family of evaluation criteria, ${g_j}:A \cup {A^R} \to \mathbb{R}$, and ${g_j}(a)$ denotes the performance of alternative $a$ on criterion $g_j$; without loss of generality, we assume that all criteria are of gain type, i.e., the greater ${g_j}(a)$, the more preferred $a$ on $g_j$, $j=1,...,n$.
\end{itemize}

\noindent The task consists in learning a preference model from the training samples composed of reference alternatives $a \in A^R$ and their assignments (denoted by $Cl(a) \in CL$) to determine the classification for non-reference alternatives $a' \in A$. Let us first define a~simple additive value function under the preferential independence hypothesis, and later extend such a preference model to consider the interactions among criteria. The additive value model $U(\cdot)$ aggregates the performances of each alternative $a \in A \cup A^R$ on all criteria into a comprehensive score:
\begin{equation}\label{eq-1}
U(a) = \sum\nolimits_{j = 1}^n {{u_{g_j}}\left( {{g_j}(a)} \right)},
\end{equation}
where $U(a)$ is a comprehensive value of $a$, and ${{u_{g_j}}\left( {{g_j}(a)} \right)}$ is a~marginal value on criterion $g_j$, $j=1,...,n$.

Since the marginal value functions $u_{g_j}(\cdot)$ for each criterion $g_j$, $j=1,...,n$, are unknown, we employ piecewise-linear marginal value functions to approximate the actual ones. Such an estimation technique has been adopted in many ordinal regression problems. Specifically, let ${X_j} = \left[ {{\alpha _j},{\beta _j}} \right]$ be the performance scale of $g_j$, such that ${{\alpha _j}}$ and ${{\beta _j}}$ are the worst and best performances, respectively. To define a~piecewise-linear marginal value function $u_{g_j}(\cdot)$, we divide ${X_j} = \left[ {{\alpha _j},{\beta _j}} \right]$ into ${\gamma _j} \geqslant 1$ equal sub-intervals, denoted by $\left[ {x_j^0,x_j^1} \right],\left[ {x_j^1,x_j^2} \right],...,\left[ {x_j^{{\gamma _j} - 1},x_j^{{\gamma _j}}} \right]$, where $x_j^k = {\alpha _j} + \frac{k}
{{{\gamma _j}}}\left( {{\beta _j} - {\alpha _j}} \right)$, $0,1,...,{\gamma _j}$. Then, the marginal value of alternative $a$ on criterion $g_j$ can be estimated through linear interpolation:
\begin{equation*}
{u_{g_j}}({g_j}(a)) = {u_{g_j}}(x_j^{{k_j}}) + \frac{{{g_j}(a) - x_j^{{k_j}}}}
{{x_j^{{k_j} + 1} - x_j^{{k_j}}}}\left( {{u_{g_j}}(x_j^{{k_j} + 1}) - {u_{g_j}}(x_j^{{k_j}})} \right),\;\;{\text{for}}\;\;{g_j}(a) \in \left[ {x_j^{{k_j}},x_j^{{k_j} + 1}} \right].
\end{equation*}
One can observe that the piecewise-linear marginal value function $u_{g_j}(\cdot)$ is fully determined by the marginal values at characteristic points, i.e., ${u_{g_j}}(x_j^0) = {u_{g_j}}({\alpha _j}),\;{u_{g_j}}(x_j^1),\;...,\;{u_{g_j}}(x_j^{{\gamma _j}}) = {u_{g_j}}({\beta _j})$. Given a sufficient number of characteristic points, the piecewise-linear marginal value function $u_{g_j}(\cdot)$ can approximate any form of non-linear value function. 

When assuming $\Delta u_{g_j}^t = {u_{g_j}}\left( {x_j^t} \right) - {u_{g_j}}\left( {x_j^{t - 1}} \right)$, $t = 1,...,{\gamma _j}$, ${u_{g_j}}({g_j}(a))$ can be rewritten as:
\begin{equation*}
{u_{g_j}}({g_j}(a)) = \sum\nolimits_{t = 1}^{{k_j}} {\Delta u_{g_j}^t}  + \frac{{{g_j}(a) - x_j^{{k_j}}}}
{{x_j^{{k_j} + 1} - x_j^{{k_j}}}}\Delta u_{g_j}^{{k_j} + 1},\;\;{\text{for}}\;{g_j}(a) \in \left[ {x_j^{{k_j}},x_j^{{k_j} + 1}} \right].
\end{equation*}
Having gathered all $\Delta u_{g_j}^t$, $t = 1,...,{\gamma _j}$ as a vector ${{\mathbf{u}}_{{g_j}}} = {\left( {\Delta u_{{g_j}}^1,...,\Delta u_{{g_j}}^{{\gamma _j}}} \right)^{\text{T}}}$ for criterion $g_j$, $j=1,...,n$, for each alternative $a \in A \cup {A^R}$ and each criterion $g_j$, $j=1,...,n$, we can define a vector ${{\mathbf{V}}_{{g_j}}}\left( a \right) = {\left( {v_{{g_j}}^1\left( a \right),...,v_{{g_j}}^{{\gamma _j}}\left( a \right)} \right)^{\text{T}}}$, such that, for each $t = 1,...,{\gamma _j}$:
\begin{equation*}
	v_{{g_j}}^t\left( a \right) = \left\{ {\begin{array}{*{20}{c}}
		{1,} \hfill & {{\text{if}}\;{g_j}\left( a \right) > x_j^t,} \hfill  \\
		{\frac{{{g_j}\left( a \right) - x_j^{t - 1}}}
			{{x_j^t - x_j^{t - 1}}},} \hfill & {{\text{if}}\;x_j^{t - 1} \leqslant {g_j}\left( a \right) \leqslant x_j^t,} \hfill  \\
		{0,} \hfill & {{\text{if}}\;{g_j}\left( a \right) < x_j^{t - 1}.} \hfill  \\
		\end{array} } \right.
\end{equation*}
Then, the marginal values $u_{g_j}(\cdot)$, $j=1,...,n$, can be represented as an~inner product between vectors as follows: 
\begin{equation*}
	{u_{g_j}}\left( {{g_j}\left( a \right)} \right) = {\mathbf{u}}_{{g_j}}^{\text{T}}{{\mathbf{V}}_{{g_j}}}\left( a \right).
\end{equation*}
Subsequently, let us denote ${\mathbf{u}} = {\left( {{\mathbf{u}}_{{g_1}}^{\text{T}},...,{\mathbf{u}}_{{g_n}}^{\text{T}}} \right)^{\text{T}}}$ and ${\mathbf{V}}\left( a \right) = {\left( {{{\mathbf{V}}_{{g_1}}}{{\left( a \right)}^{\text{T}}},...,{{\mathbf{V}}_{{g_n}}}{{\left( a \right)}^{\text{T}}}} \right)^{\text{T}}}$, and the comprehensive value $U(\cdot)$ can be expressed in the following way:
\begin{equation*}
	U\left( a \right) = {{\mathbf{u}}^{\text{T}}}{\mathbf{V}}\left( a \right).
\end{equation*}

\subsection{Learning preference model from reference alternatives}

\noindent In this section, we propose a~new method for learning a preference model in form of an additive value function~(\ref{eq-1}) from a set of reference alternatives $a \in A^R$ and their associated precise class assignments $Cl(a)$. Before describing the estimation procedure, let us present the underlying consistency principle for characterizing the preference relation between alternatives in sorting problems.

\begin{description}
	\item[Definition 1.] For any pair of alternatives $a,b \in A \cup A^R$, value function $U\left(  \cdot  \right)$ is said to be \textit{consistent} with the assignments of $a$ and $b$ (denoted by $Cl(a)$ and $Cl(b)$, respectively, and $Cl(a),Cl(b) \in CL$) iff:
	\begin{equation}\label{eq-2}
	U\left( {a} \right) \geqslant U\left( {b} \right) \Rightarrow Cl\left( {a} \right) \succsim Cl\left( {b} \right),
	\end{equation}
	where $\succsim$ means ``at least as good as''.  Observe that implication (\ref{eq-2}) is equivalent to:
	\begin{equation}\label{eq-3}
	Cl\left( {a} \right) \succ Cl\left( {b} \right) \Rightarrow U\left( {a} \right) > U\left( {b} \right).
	\end{equation}
\end{description}

\noindent According to Definition 1, value function $U\left(  \cdot  \right)$ inferred from the analysis of assignment examples should ensure $U\left( {a} \right) > U\left( {b} \right)$ for pairs of reference alternatives $\left( {a,b} \right) \in {A^R} \times {A^R}$ such that $Cl\left( {a} \right) \succ Cl\left( {b} \right)$. However, there may exist no such a~value function that would guarantee perfect compatibility due to the inconsistency of some assignment examples with an assumed preference model. In turn, we can estimate a value function that would maximize the difference between $U\left( {a} \right)$ and $U\left( {b} \right)$ for pairs of reference alternatives $\left( {a,b} \right) \in {A^R} \times {A^R}$ such that $Cl\left( {a} \right) \succ Cl\left( {b} \right)$. This can be implemented by solving the following linear programming model:
\begin{align}
	({\text{P0}}): \;\;\;\;\;\; & Minimize\; -d, \label{eq-4} \\ 
	{\text{s}}{\text{.t}}{\text{.}} \;\;\;\;\;\; & U\left( a \right) - U\left( b \right) \geqslant d\left( {a,b} \right),\;\;a \in A_{s + 1}^R,\;b \in A_s^R,\;s = 1,...,q - 1, \label{eq-5} \\ 
	& \frac{1}{{\left| {A_{s + 1}^R} \right|\left| {A_s^R} \right|}}\sum\nolimits_{a \in A_{s + 1}^R,\;b \in A_s^R} {d\left( {a,b} \right)}  \geqslant d,\;\;s = 1,...,q - 1, \label{eq-6}
\end{align} 
where $A_s^R$ denotes a~set of reference alternatives assigned to class $Cl_s$, and ${\left| {A_s^R} \right|}$ is the cardinality of $A_s^R$. For any $(a,b) \in A_{s + 1}^R \times A_s^R$, $s = 1,...,q - 1$, the value difference for such a pair of alternatives, denoted by $d\left( {a,b} \right)$, is identified by constraint (\ref{eq-5}). In constraint (\ref{eq-6}), $\frac{1}
{{\left| {A_{s + 1}^R} \right|\left| {A_s^R} \right|}}\sum\nolimits_{a \in A_{s + 1}^R,\;b \in A_s^R} {d\left( {a,b} \right)}$ is the average value difference for pairs $(a,b) \in A_{s + 1}^R \times A_s^R$. Then, $d$ corresponds to the minimum of such value differences for all consecutive classes. By maximizing the minimum value difference $d$ (i.e., $Minimize\; -d$), model (P0) aims to find value function $U\left(  \cdot  \right)$ that restores the assignment consistency as accurately as possible.

Note that it would not be appropriate to maximize the minimal value difference between reference alternatives from the consecutive classes (i.e., replace constraint (\ref{eq-6}) with constraint $d\left( {a,b} \right) \geqslant d$, $a \in A_{s + 1}^R$, $b \in A_s^R$, $s = 1,...,q - 1$). In case of an inconsistent reference set, such a minimal value is smaller than zero (i.e., $d < 0$), and then it is meaningless to maximize it. Moreover, we do not maximize the sum of value differences between reference alternatives from the consecutive classes (i.e., remove constraint (\ref{eq-6}) and replace objective (\ref{eq-4}) with $Minimize\; - \sum\limits_{s = 1,...,q - 1} {\sum\limits_{a \in A_{s + 1}^R,b \in A_s^R} {d\left( {a,b} \right)} }$). The underlying reason for avoiding doing so can be illustrated through a simple example: let us consider three reference alternatives $a \in A_{s + 1}^R$, $b \in A_s^R$ and $c \in A_{s - 1}^R$. Then, $d\left( {a,b} \right) = U\left( a \right) - U\left( b \right)$ and $d\left( {b,c} \right) = U\left( b \right) - U\left( c \right)$. If we aimed to maximize the sum of value differences between reference alternatives from the consecutive classes, we would just maximize the difference between $U\left( a \right)$ and $U\left( c \right)$, because $d\left( {a,b} \right) + d\left( {b,c} \right) = (U\left( a \right) - U\left( b \right)) + (U\left( b \right) - U\left( c \right)) = U\left( a \right) - U\left( c \right)$ and then $d\left( {a,b} \right)$ and $d\left( {b,c} \right)$ would be completely neglected.

When addressing a large number of reference alternatives $A^R$, model (P0) contains a huge number of constrains (\ref{eq-5}), exceeding the processing capacity of existing linear programming solvers. Thus, we propose a~method for transforming model (P0) to an equivalent model that is suitable for data-intensive tasks. The key idea is to aggregate constrains (\ref{eq-5}) for all the possible pairs of reference alternatives $(a,b) \in A_{s + 1}^R \times A_s^R$ for a~particular $s \in \left\{ {1,...,q - 1} \right\}$, and obtain:
\begin{equation}\label{eq-7}
	\left| {A_s^R} \right|\sum\nolimits_{a \in A_{s + 1}^R} {U\left( a \right)}  - \left| {A_{s + 1}^R} \right|\sum\nolimits_{b \in A_s^R} {U\left( b \right)}  \geqslant \sum\nolimits_{a \in A_{s + 1}^R,\;b \in A_s^R} {d\left( {a,b} \right)}.
\end{equation}
By dividing ${\left| {A_{s + 1}^R} \right|\left| {A_s^R} \right|}$, constraint (\ref{eq-7}) is equivalent to:
\begin{equation}\label{eq-8}
	\frac{1}
	{{\left| {A_{s + 1}^R} \right|}}\sum\nolimits_{a \in A_{s + 1}^R} {U\left( a \right)}  - \frac{1}
	{{\left| {A_s^R} \right|}}\sum\nolimits_{b \in A_s^R} {U\left( b \right)}  \geqslant \frac{1}
	{{\left| {A_{s + 1}^R} \right|\left| {A_s^R} \right|}}\sum\nolimits_{a \in A_{s + 1}^R,\;b \in A_s^R} {d\left( {a,b} \right)}.
\end{equation}
Then, since $U\left( a \right) = {{\mathbf{u}}^{\text{T}}}{\mathbf{V}}\left( a \right)$, constraint (\ref{eq-8}) can be transformed to:
\begin{equation}\label{eq-9}
	{{\mathbf{u}}^{\text{T}}}\left( {\frac{1}
		{{\left| {A_{s + 1}^R} \right|}}\sum\nolimits_{a \in A_{s + 1}^R} {{\mathbf{V}}\left( a \right)} } \right) - {{\mathbf{u}}^{\text{T}}}\left( {\frac{1}
		{{\left| {A_s^R} \right|}}\sum\nolimits_{b \in A_s^R} {{\mathbf{V}}\left( b \right)} } \right) \geqslant \frac{1}
	{{\left| {A_{s + 1}^R} \right|\left| {A_s^R} \right|}}\sum\nolimits_{a \in A_{s + 1}^R,\;b \in A_s^R} {d\left( {a,b} \right)}.
\end{equation} 
For each class $Cl_s$, $s=1,...,q$, let us average ${{\mathbf{V}}\left( a \right)}$ for all ${a \in A_s^R}$ and derive:
\begin{equation}\label{eq-10}
	{{\boldsymbol{\mu }}_s} = \frac{1}
	{{\left| {A_s^R} \right|}}\sum\nolimits_{a \in A_s^R} {{\mathbf{V}}\left( a \right)}.
\end{equation}
Then, constraint (\ref{eq-10}) can be written as:
\begin{equation} \label{eq-11}
	{{\mathbf{u}}^{\text{T}}}{{\boldsymbol{\mu }}_{s + 1}} - {{\mathbf{u}}^{\text{T}}}{{\boldsymbol{\mu }}_s} \geqslant \frac{1}
	{{\left| {A_{s + 1}^R} \right|\left| {A_s^R} \right|}}\sum\nolimits_{a \in A_{s + 1}^R,\;b \in A_s^R} {d\left( {a,b} \right)}.
\end{equation}
In this way, model (P0) can be reformulated as:
\begin{align}
({\text{P1}}): \;\;\;\;\;\; & Minimize\; -d, \label{eq-12} \\ 
{\text{s}}{\text{.t}}{\text{.}} \;\;\;\;\;\; & {{\mathbf{u}}^{\text{T}}}{{\boldsymbol{\mu }}_{s + 1}} - {{\mathbf{u}}^{\text{T}}}{{\boldsymbol{\mu }}_s} \geqslant d,\;\;s = 1,...,q - 1. \label{eq-13}
\end{align} 
Note that the number of constraints (\ref{eq-13}) is related only to the number of classes (i.e., $q-1$) rather than the number of pairs of reference alternatives. Thus, model (P1) can deal with a~large set of reference alternatives efficiently.

\begin{thm}
	The optimal value of the objective function of model (P0) is equal to that of model (P1).
\end{thm}
\begin{description}
	\item[Proof.] See Appendix A. \qed
\end{description}

\noindent In addition to maximizing the value difference for pairs of reference alternatives from the consecutive classes, we also propose to minimize the value difference among reference alternatives from the same class. In this way, the distribution of comprehensive values of reference alternatives from the same class would be more concentrated, and consecutive classes could be clearly delimited. For this purpose, we aim at minimizing the second objective:
\begin{equation*}
	\begin{gathered}
	\sum\limits_{i = 1}^q {\sum\limits_{a,b \in A_i^R} {{{\left( {U\left( a \right) - U\left( b \right)} \right)}^2}} } 
	= \sum\limits_{i = 1}^q {\sum\limits_{a,b \in A_i^R} {{{\left( {{{\mathbf{u}}^{\text{T}}}{\mathbf{v}}\left( a \right) - {{\mathbf{u}}^{\text{T}}}{\mathbf{v}}\left( b \right)} \right)}^2}} } 
	= \sum\limits_{i = 1}^q {\sum\limits_{a,b \in A_i^R} {{{\mathbf{u}}^{\text{T}}}\left( {{\mathbf{v}}\left( a \right) - {\mathbf{v}}\left( b \right)} \right){{\left( {{\mathbf{v}}\left( a \right) - {\mathbf{v}}\left( b \right)} \right)}^{\text{T}}}{\mathbf{u}}} }  \hfill \\
	= {{\mathbf{u}}^{\text{T}}}\left( {\sum\limits_{i = 1}^q {\sum\limits_{a,b \in A_i^R} {\left( {{\mathbf{v}}\left( a \right) - {\mathbf{v}}\left( b \right)} \right){{\left( {{\mathbf{v}}\left( a \right) - {\mathbf{v}}\left( b \right)} \right)}^{\text{T}}}} } } \right){\mathbf{u}} 
	= {{\mathbf{u}}^{\text{T}}}{\mathbf{Su}}, \hfill \\ 
	\end{gathered}
\end{equation*}
where ${\mathbf{S}} = {\sum\limits_{i = 1}^q {\sum\limits_{a,b \in A_i^R} {\left( {{\mathbf{v}}\left( a \right) - {\mathbf{v}}\left( b \right)} \right){{\left( {{\mathbf{v}}\left( a \right) - {\mathbf{v}}\left( b \right)} \right)}^{\text{T}}}} } }$ is a $\gamma  \times \gamma$ matrix and $\gamma  = \sum\limits_{j = 1}^n {{\gamma _j}}$. Since $\sum\limits_{i = 1}^q {\sum\limits_{a,b \in A_i^R} {{{\left( {U\left( a \right) - U\left( b \right)} \right)}^2}} }  \geqslant 0$ always holds, ${\mathbf{S}}$ must be positive semi-definite. Then, putting together the above two objectives, we propose the following convex quadratic programming model:
\begin{align}
({\text{P2}}): \;\;\;\;\;\; & \min  - d + {C_1}{{\mathbf{u}}^{\text{T}}}{\mathbf{Su}} + {C_2}\left\| {\mathbf{u}} \right\|_2^2, \label{eq-14} \\ 
{\text{s}}{\text{.t}}{\text{.}} \;\;\;\;\;\; & {{\mathbf{u}}^{\text{T}}}{{\boldsymbol{\mu }}_{k + 1}} \geqslant {{\mathbf{u}}^{\text{T}}}{{\boldsymbol{\mu }}_k} + d,\;\;k = 1,...,q - 1, \label{eq-15} \\
& {{\mathbf{u}}^{\text{T}}}{{\mathbf{V}}^*} = 1, \label{eq-16}\\
& d \geqslant 0, \label{eq-17}\\
& {\mathbf{u}} \geqslant {\mathbf{0}}, \label{eq-18}
\end{align} 
where ${{\mathbf{V}}^*}$ is a $\gamma$-dimensional vector with all entries being equal to one, and constraint (\ref{eq-16}) is used to bound ${U\left( \cdot \right)}$ to the interval [0,1]. Since class $Cl_{k+1}$ is preferred to class $Cl_k$, $k=1,...,q-1$, we require that $d \geqslant 0$ in constraint (\ref{eq-17}). Besides the two aforementioned objectives, the regularization term $\left\| {\mathbf{u}} \right\|_2^2$ is added to the objective of model (P2) to avoid the problem of over-fitting. Specifically, since the performance scale on each criterion is divided into a certain number of equal sub-intervals, the fitting ability of the estimated function improves with the increase in the number of sub-intervals, at the same time increasing the risk of over-fitting. The regularization term $\left\| {\mathbf{u}} \right\|_2^2$, also named Tikhonov regularization, penalizes functions that are ``too wiggly", and derives marginal value functions that are as ``smooth" as possible, which alleviates the problem of over-fitting caused by inappropriately dividing the performance scale into too many sub-intervals. The constants ${C_1},{C_2} > 0$ are used to make a~trade-off between the two objectives and the regularization term. Values of ${C_1}$ and ${C_2}$ can be chosen through \emph{$K$-fold cross-validation} in the following manner: the whole set of reference alternatives $A^R$ is randomly partitioned into $K$ (usually $K$ is set to be 5 or 10) equal sized folds such that the percentage of reference alternatives from different decision classes in each fold are the same with that in $A^R$. For certain $C_1$ and $C_2$, $K-1$ folds are used as the training data and the remaining fold is retained as the validation data for testing the model. The cross-validation process is repeated $K$ times, and then the $K$ results are averaged to evaluate the performance of the developed model (e.g., classification accuracy). Finally, the values of $C_1$ and $C_2$ corresponding to the best performance are chosen as the optimal setting for the two parameters. For model (P2), let us remark that, since $\mathbf{S}$ and $\boldsymbol{\mu}_k$, $k=1,...,q$, can be specified in advance, model (P2) is not related to the pairwise comparisons among reference alternatives, and thus the number of constraints is small. Hence, model (P2) can be easily solved using popular optimization packages, such as Lingo, Cplex, or MATLAB.

\subsection{Considering interactions among criteria}
\noindent Even though an additive value function model is widely used in real-world decision aiding, it is not able to represent interactions among criteria due to the underlying preferential independence hypothesis. To handle interactions among criteria, we incorporate and adjust the model so that to propose a new method which can address a large set of reference alternatives efficiently. The underlying model is an additive value function augmented by ``bonus" and ``penalty'" components for, respectively, positive and negative interactions among criteria, which is formulated as:
\begin{equation}\label{eq-19}
	U\left( a \right) = \sum\limits_{j = 1}^n {{u_{g_j}}\left( {{g_j}\left( a \right)} \right)}  + \sum\limits_{\left\{ {{g_j},{g_k}} \right\} \in G:j < k} {syn_{{g_j},{g_k}}^ + \left( {{g_j}\left( a \right),{g_k}\left( a \right)} \right)}  - \sum\limits_{\left\{ {{g_j},{g_k}} \right\} \in G:j < k} {syn_{{g_j},{g_k}}^ - \left( {{g_j}\left( a \right),{g_k}\left( a \right)} \right)},
\end{equation}
where $syn_{{g_j},{g_k}}^ + \left( { \cdot , \cdot } \right)$ and $syn_{{g_j},{g_k}}^ - \left( { \cdot , \cdot } \right)$ are the bonus and penalty values for modelling the interactions between $g_j$ and $g_k$. The extended form (\ref{eq-19}) of value function should fulfil the following basic conditions:
\begin{itemize}
	\item normalization: $U\left( a \right) = 0$ if ${g_j}\left( a \right) = {\alpha _j}$, $j = 1,...,n$, and $U\left( a \right) = 1$ if ${g_j}\left( a \right) = {\beta _j}$, $j = 1,...,n$,
	\item monotonicity (a): $\forall \left\{ {{g_j},{g_k}} \right\} \in G, j<k$, if ${g_j}\left( a \right) \geqslant {g_j}\left( b \right)$ and ${g_k}\left( a \right) \geqslant {g_k}\left( b \right)$, then $syn_{{g_j},{g_k}}^ + \left( {{g_j}\left( a \right),{g_k}\left( a \right)} \right) \geqslant syn_{{g_j},{g_k}}^ + \left( {{g_j}\left( b \right),{g_k}\left( b \right)} \right)$ and $syn_{{g_j},{g_k}}^ - \left( {{g_j}\left( a \right),{g_k}\left( a \right)} \right) \geqslant syn_{{g_j},{g_k}}^ - \left( {{g_j}\left( b \right),{g_k}\left( b \right)} \right)$,
	\item monotonicity (b): $\forall H \subseteq G$, if ${g_j}\left( a \right) \geqslant {g_j}\left( b \right)$, $\forall {g_j} \in H$, then:
	\begin{equation*}
		\begin{gathered}
		\sum\limits_{{g_j} \in H} {{u_{g_j}}\left( {{g_j}\left( a \right)} \right)}  + \sum\limits_{\left\{ {{g_j},{g_k}} \right\} \in H:j < k} {syn_{{g_j},{g_k}}^ + \left( {{g_j}\left( a \right),{g_k}\left( a \right)} \right)}  - \sum\limits_{\left\{ {{g_j},{g_k}} \right\} \in H:j < k} {syn_{{g_j},{g_k}}^ - \left( {{g_j}\left( a \right),{g_k}\left( a \right)} \right)}  \hfill \\
		\geqslant \sum\limits_{{g_j} \in H} {{u_{g_j}}\left( {{g_j}\left( b \right)} \right)}  + \sum\limits_{\left\{ {{g_j},{g_k}} \right\} \in H:j < k} {syn_{{g_j},{g_k}}^ + \left( {{g_j}\left( b \right),{g_k}\left( b \right)} \right)}  - \sum\limits_{\left\{ {{g_j},{g_k}} \right\} \in H:j < k} {syn_{{g_j},{g_k}}^ - \left( {{g_j}\left( b \right),{g_k}\left( b \right)} \right)}.  \hfill \\ 
		\end{gathered}
	\end{equation*}
\end{itemize}
The normalization conditions require ${U\left( \cdot \right)}$ to be bounded to the interval [0,1]. Monotonicity (a) ensures that $syn_{{g_j},{g_k}}^ + \left( { \cdot , \cdot } \right)$ and $syn_{{g_j},{g_k}}^ - \left( { \cdot , \cdot } \right)$ are monotone non-decreasing with respect to their arguments. Monotonicity (b) can be interpreted as follows: when comparing any pair of alternatives $a,b$ on a~subset of criteria $H \subseteq G$, if $a$~is at least as good as $b$ for all ${g_j} \in H$, the comprehensive value of $a$ derived from the analysis of $H$ should be not worse than that of $b$. Note that monotonicity (b) induces numerous constraints as $G$ has ${2^n} - 1$ non-empty subsets. For this reason, we assume that any criterion interacts with at most one other. This makes both the inference of a value function more tractable and the constructed model more interpretable. Under this assumption, value function (\ref{eq-19}) can be reformulated as:
\begin{equation}\label{eq-20}
	\begin{gathered}
	U\left( a \right) = \sum\limits_{{g_i} \notin { \cup _{\left\{ {{g_j},{g_k}} \right\} \in Syn}}\left\{ {{g_j},{g_k}} \right\}} {{u_{g_i}}\left( {{g_i}\left( a \right)} \right)}  \hfill \\
	\;\;\;\;\;\;\;\;\;\;\;\; + \sum\limits_{\left\{ {{g_j},{g_k}} \right\} \in Syn} {\left( {{u_{g_j}}\left( {{g_j}\left( a \right)} \right) + {u_k}\left( {{g_k}\left( a \right)} \right) + syn_{{g_j},{g_k}}^ + \left( {{g_j}\left( a \right),{g_k}\left( a \right)} \right) - syn_{{g_j},{g_k}}^ - \left( {{g_j}\left( a \right),{g_k}\left( a \right)} \right)} \right)},  \hfill \\ 
	\end{gathered}
\end{equation}
where
\begin{equation*}
	\begin{gathered}
	Syn = \left\{ {\left\{ {{g_j},{g_k}} \right\} \subseteq G,\;j < k\; : \;syn_{{g_j},{g_k}}^ + \left( {{g_j}\left( a \right),{g_k}\left( a \right)} \right) \ne 0\;\;\;{\text{or}}} \right.
	\;\;\left. {syn_{{g_j},{g_k}}^ - \left( {{g_j}\left( a \right),{g_k}\left( a \right)} \right) \ne 0\;{\text{for}}\;{\text{some}}\;a} \right\} \hfill \\ 
	\end{gathered}
\end{equation*}
denotes the set of all pairs ${\left\{ {{g_j},{g_k}} \right\}}$ of interacting criteria. Value function (\ref{eq-20}) divides a~set of criteria into two disjoint subsets: one consisting of criteria not interacting with the remaining ones, and the other composed of the interacting criteria. In this way, monotonicity (b) can be reduced to that, $\forall \left\{ {{g_j},{g_k}} \right\} \in Syn$, if ${g_j}\left( a \right) \geqslant {g_j}\left( b \right)$ and ${g_k}\left( a \right) \geqslant {g_k}\left( b \right)$, then:
\begin{equation}\label{eq-21}
	\begin{gathered}
	{u_{g_j}}\left( {{g_j}\left( a \right)} \right) + {u_k}\left( {{g_k}\left( a \right)} \right) + syn_{{g_j},{g_k}}^ + \left( {{g_j}\left( a \right),{g_k}\left( a \right)} \right) - syn_{{g_j},{g_k}}^ - \left( {{g_j}\left( a \right),{g_k}\left( a \right)} \right) \hfill \\
	\;\;\;\; \geqslant {u_{g_j}}\left( {{g_j}\left( b \right)} \right) + {u_k}\left( {{g_k}\left( b \right)} \right) + syn_{{g_j},{g_k}}^ + \left( {{g_j}\left( b \right),{g_k}\left( b \right)} \right) - syn_{{g_j},{g_k}}^ - \left( {{g_j}\left( b \right),{g_k}\left( b \right)} \right). \hfill \\ 
	\end{gathered}
\end{equation}

\noindent In this study, we propose to define the bonus and penalty components $syn_{{g_j},{g_k}}^ + \left( { \cdot , \cdot } \right)$ and $syn_{{g_j},{g_k}}^ - \left( { \cdot , \cdot } \right)$ in the following way:
\begin{align}
	syn_{{g_j},{g_k}}^ + \left( {{g_j}\left( a \right),{g_k}\left( a \right)} \right) = \sum\limits_{s = 1}^{{\gamma _j}} {\sum\limits_{t = 1}^{{\gamma _k}} {\eta _{{g_j},{g_k}}^{ + ,s,t}v_{{g_j}}^s\left( a \right)v_{{g_k}}^t\left( a \right)} }, \label{eq-22}\\
	syn_{{g_j},{g_k}}^ - \left( {{g_j}\left( a \right),{g_k}\left( a \right)} \right) = \sum\limits_{s = 1}^{{\gamma _j}} {\sum\limits_{t = 1}^{{\gamma _k}} {\eta _{{g_j},{g_k}}^{ - ,s,t}v_{{g_j}}^s\left( a \right)v_{{g_k}}^t\left( a \right)} }, \label{eq-23}
\end{align}
where $\eta _{{g_j},{g_k}}^{ + ,s,t},\eta _{{g_j},{g_k}}^{ - ,s,t} \geqslant 0$ are the coefficients for modelling the positive and negative interactions on criterion $g_j$'s $s$-th sub-interval and criterion $g_k$'s $t$-th sub-interval. {Since the above definition of $syn_{{g_j},{g_k}}^ + \left( {{g_j}\left( a \right),{g_k}\left( a \right)} \right)$ and $syn_{{g_j},{g_k}}^ - \left( {{g_j}\left( a \right),{g_k}\left( a \right)} \right)$ is based on the product of ${v_{{g_j}}^s\left( a \right)}$ and ${v_{{g_k}}^t\left( a \right)}$, it ensures that the bonus and penalty components are monotone and strictly increasing with respect to ${g_j} \left( a \right)$ and ${g_k} \left( a \right)$ for any pair of criteria~$(g_j,g_k)$. 
	
An alternative definition of the interaction components would consist in deriving the minimum from ${v_{{g_j}}^s\left( a \right)}$ and ${v_{{g_k}}^t\left( a \right)}$ as follows:
\begin{align}
	syn_{{g_j},{g_k}}^ + \left( {{g_j}\left( a \right),{g_k}\left( a \right)} \right) = \sum\nolimits_{s = 1}^{{\gamma _j}} {\sum\nolimits_{t = 1}^{{\gamma _k}} {\eta _{{g_j},{g_k}}^{ + ,s,t}\min \left\{ {v_{{g_j}}^s\left( a \right),v_{{g_k}}^t\left( a \right)} \right\}} }, \label{eq-22-1}\\
	syn_{{g_j},{g_k}}^ - \left( {{g_j}\left( a \right),{g_k}\left( a \right)} \right) = \sum\nolimits_{s = 1}^{{\gamma _j}} {\sum\nolimits_{t = 1}^{{\gamma _k}} {\eta _{{g_j},{g_k}}^{ - ,s,t}\min \left\{ {v_{{g_j}}^s\left( a \right),v_{{g_k}}^t\left( a \right)} \right\}} }. \label{eq-23-1}
\end{align}
This makes these components monotone non-decreasing with respect to ${g_j} \left( a \right)$ and ${g_k} \left( a \right)$.} It is easy to validate that such two types of definitions of $syn_{{g_j},{g_k}}^ + \left( { \cdot , \cdot } \right)$ and $syn_{{g_j},{g_k}}^ - \left( { \cdot , \cdot } \right)$ satisfy monotonicity (a). To ensure monotonicity (b), let us consider the following proposition.

\begin{thm}
	Value function (\ref{eq-20}), in which the bonus and penalty components are, respectively, defined as (\ref{eq-22}) and (\ref{eq-23}) (or (\ref{eq-22-1}) and (\ref{eq-23-1})), satisfies monotonicity (b), if and only if $\forall \left\{ {{g_j},{g_k}} \right\} \in Syn$:
	\begin{equation}\label{eq-24}
	\Delta u_{{g_j}}^s + \sum\limits_{q = 1}^t {\left( {\eta _{{g_j},{g_k}}^{ + ,s,q} - \eta _{{g_j},{g_k}}^{ - ,s,q}} \right)}  \geqslant 0,\;\;s = 1,...,{\gamma _j},\;t = 1,...,{\gamma _k}.
	\end{equation}
\end{thm}
\begin{description}
	\item[Proof.] See Appendix B. \qed
\end{description}

\noindent Let us now introduce the method for estimating the coefficients $\eta _{{g_j},{g_k}}^{ + ,s,t}$ and $\eta _{{g_j},{g_k}}^{ - ,s,t}$, $s = 1,...,{\gamma _j}$, $t = 1,...,{\gamma _k}$, ${g_j},{g_k} \in G,\;j < k$, from the assignments of reference alternatives. For the convenience of the analysis, let us define the following vectors:
\begin{equation*}
	\begin{gathered}
	{\mathbf{V}}_{{g_j},{g_k}}^ + \left( a \right) = \left( {v_{{g_j}}^1\left( a \right)v_{{g_k}}^1\left( a \right),...,v_{{g_j}}^1\left( a \right)v_{{g_k}}^{{\gamma _k}}\left( a \right),} \right. \ldots, 
	{\left. {v_{{g_j}}^{{\gamma _j}}\left( a \right)v_{{g_k}}^1\left( a \right),...,v_{{g_j}}^{{\gamma _j}}\left( a \right)v_{{g_k}}^{{\gamma _k}}\left( a \right)} \right)^{\text{T}}}, \hfill \\ 
	\end{gathered}
\end{equation*}
\begin{equation*}
	\begin{gathered}
	{\mathbf{V}}_{{g_j},{g_k}}^ - \left( a \right) = \left( { - v_{{g_j}}^1\left( a \right)v_{{g_k}}^1\left( a \right),..., - v_{{g_j}}^1\left( a \right)v_{{g_k}}^{{\gamma _k}}\left( a \right),} \right.
	\ldots ,
	{\left. {- v_{{g_j}}^{{\gamma _j}}\left( a \right)v_{{g_k}}^1\left( a \right),..., - v_{{g_j}}^{{\gamma _j}}\left( a \right)v_{{g_k}}^{{\gamma _k}}\left( a \right)} \right)^{\text{T}}}, \hfill \\ 
	\end{gathered}
\end{equation*}
\begin{equation*}
	{\boldsymbol{\eta }}_{{g_j},{g_k}}^ +  = {\left( {\eta _{{g_j},{g_k}}^{ + ,1,1},...,\eta _{{g_j},{g_k}}^{ + ,1,{\gamma _k}},\ldots,\eta _{{g_j},{g_k}}^{ + ,{\gamma _j},1},...,\eta _{{g_j},{g_k}}^{ + ,{\gamma _j},{\gamma _k}}} \right)^{\text{T}}},
\end{equation*}
\begin{equation*}
	{\boldsymbol{\eta }}_{{g_j},{g_k}}^ -  = {\left( {\eta _{{g_j},{g_k}}^{ - ,1,1},...,\eta _{{g_j},{g_k}}^{ - ,1,{\gamma _k}},\ldots ,\eta _{{g_j},{g_k}}^{ - ,{\gamma _j},1},...,\eta _{{g_j},{g_k}}^{ - ,{\gamma _j},{\gamma _k}}} \right)^{\text{T}}}.
\end{equation*}
Then, the bonus and penalty components $syn_{{g_j},{g_k}}^ + \left( { \cdot , \cdot } \right)$ and $syn_{{g_j},{g_k}}^ - \left( { \cdot , \cdot } \right)$ can be reformulated in the form of an~inner product between the above vectors as follows:
\begin{equation*}
	syn_{{g_j},{g_k}}^ + \left( {{g_j}\left( a \right),{g_k}\left( a \right)} \right) = {{\boldsymbol{\eta }}{_{{g_j},{g_k}}^ +} ^{\text{T}}}{\mathbf{V}}_{{g_j},{g_k}}^ + \left( a \right),
\end{equation*}
\begin{equation*}
	syn_{{g_j},{g_k}}^ - \left( {{g_j}\left( a \right),{g_k}\left( a \right)} \right) =  - {{\boldsymbol{\eta }}{_{{g_j},{g_k}}^ -} ^{\text{T}}}{\mathbf{V}}_{{g_j},{g_k}}^ - \left( a \right).
\end{equation*}
By considering the interactions between criteria, we can redefine ${\mathbf{V}}\left( a \right)$ and ${\mathbf{u}}$ as follows:
\begin{equation*}
	\begin{gathered}
	{\mathbf{V}^\text{INT}}\left( a \right) = \left( {{{\mathbf{V}}_{{g_1}}}{{\left( a \right)}^{\text{T}}},...,{{\mathbf{V}}_{{g_n}}}{{\left( a \right)}^{\text{T}}},} \right. \hfill \\
	\;\;\;\;\;\;\;\;\;\;\;\;{\mathbf{V}}_{{g_1},{g_2}}^ + {\left( a \right)^{\text{T}}},{\mathbf{V}}_{{g_1},{g_2}}^ - {\left( a \right)^{\text{T}}},...,{\mathbf{V}}_{{g_1},{g_n}}^ + {\left( a \right)^{\text{T}}},{\mathbf{V}}_{{g_1},{g_n}}^ - {\left( a \right)^{\text{T}}}, \hfill \\
	\;\;\;\;\;\;\;\;\;\;\;\;{\mathbf{V}}_{{g_2},{g_3}}^ + {\left( a \right)^{\text{T}}},{\mathbf{V}}_{{g_2},{g_3}}^ - {\left( a \right)^{\text{T}}},...,{\mathbf{V}}_{{g_2},{g_n}}^ + {\left( a \right)^{\text{T}}},{\mathbf{V}}_{{g_2},{g_n}}^ - {\left( a \right)^{\text{T}}}, \hfill \\
	\;\;\;\;\;\;\;\;\;\;\;\;......, \hfill \\
	\;\;\;\;\;\;\;\;\;\;\;\;{\left. {{\mathbf{V}}_{{g_{n - 1}},{g_n}}^ + {{\left( a \right)}^{\text{T}}},{\mathbf{V}}_{{g_{n - 1}},{g_n}}^ - {{\left( a \right)}^{\text{T}}}} \right)^{\text{T}}}, \hfill \\ 
	\end{gathered}
\end{equation*}
\begin{equation*}
	\begin{gathered}
	{\mathbf{u}^\text{INT}} = \left( {{\mathbf{u}}_{{g_1}}^{\text{T}},...,{\mathbf{u}}_{{g_n}}^{\text{T}},} \right. \hfill \\
	\;\;\;\;\;\;\;{\boldsymbol{\eta }}{_{{g_1},{g_2}}^ +} ^{\text{T}},{\boldsymbol{\eta }}{_{{g_1},{g_2}}^ -} ^{\text{T}},...,{\boldsymbol{\eta }}{_{{g_1},{g_n}}^ +} ^{\text{T}},{\boldsymbol{\eta }}{_{{g_1},{g_n}}^ -} ^{\text{T}}, \hfill \\
	\;\;\;\;\;\;\;{\boldsymbol{\eta }}{_{{g_2},{g_3}}^ +} ^{\text{T}},{\boldsymbol{\eta }}{_{{g_2},{g_3}}^ -} ^{\text{T}},...,{\boldsymbol{\eta }}{_{{g_2},{g_n}}^ +} ^{\text{T}},{\boldsymbol{\eta }}{_{{g_2},{g_n}}^ -} ^{\text{T}}, \hfill \\
	\;\;\;\;\;\;\;......, \hfill \\
	\;\;\;\;\;\;\;{\left. {{\boldsymbol{\eta }}{{_{{g_{n - 1}},{g_n}}^ + }^{\text{T}}},{\boldsymbol{\eta }}{{_{{g_{n - 1}},{g_n}}^ - }^{\text{T}}}} \right)^{\text{T}}}, \hfill \\ 
	\end{gathered}
\end{equation*}
where ${\mathbf{V}^\text{INT}}\left( a \right)$ and ${\mathbf{u}^\text{INT}}$ are constructed by adding more dimensions for characterizing interactions between criteria. In this way, according to (\ref{eq-19}), the comprehensive score of alternative $a$ can be formulated as:
\begin{equation}\label{eq-a2}
	U(a) = {{\mathbf{u}^\text{INT}}^{\text{T}}}{\mathbf{V}^\text{INT}}(a),
\end{equation}
which is consistent with the form of an~additive value function under the preferential independence hypothesis.

Utilizing Proposition 2, we can replace the value function in model (P2) with (\ref{eq-19}) and incorporate additional constraints related to modelling the interactions between criteria. The optimization model adapted from (P2) for considering the interactions among criteria can be formulated in the following way:

\begin{align}
({\text{P3}}): \;\;\;\;\;\; & Minimize\;  - d + {C_1}{{\mathbf{u}^\text{INT}}^{\text{T}}}{\mathbf{S}^\text{INT}\mathbf{u}^\text{INT}} + {C_2}\left\| {\mathbf{u}^\text{INT}} \right\|_2^2, \label{eq-25} \\ 
{\text{s}}{\text{.t}}{\text{.}} \;\;\;\;\;\; & {{\mathbf{u}^\text{INT}}^{\text{T}}}{{\boldsymbol{\mu }}_{k + 1}^\text{INT}} \geqslant {{\mathbf{u}^\text{INT}}^{\text{T}}}{{\boldsymbol{\mu }}_k^\text{INT}} + d,\;\;k = 1,...,q - 1, \label{eq-26} \\
& \Delta u_{{g_j}}^s + \sum\limits_{q = 1}^t {\left( {\eta _{{g_j},{g_k}}^{ + ,s,q} - \eta _{{g_j},{g_k}}^{ - ,s,q}} \right)}  \geqslant 0,\;\;s = 1,...,{\gamma _j},\;\;t = 1,...,{\gamma _k},\;\;{g_j},{g_k} \in G,\;\;j < k, \label{eq-27}\\
& \eta _{{g_j},{g_k}}^{ + ,s,t} \leqslant \chi _{{g_j},{g_k}}^ + ,\;\;\eta _{{g_j},{g_k}}^{ - ,s,t} \leqslant \chi _{{g_j},{g_k}}^ - ,\;\;s = 1,...,{\gamma _j},\;\;t = 1,...,{\gamma _k},\;\;{g_j},{g_k} \in G,\;\;j < k, \label{eq-28}\\
& \chi _{{g_j},{g_k}}^ +  + \chi _{{g_j},{g_k}}^ -  \leqslant 1,\;\;{g_j},{g_k} \in G,\;\;j < k, \label{eq-29}\\
& \sum\nolimits_{{g_k} \in G} {\left( {\chi _{{g_j},{g_k}}^ +  + \chi _{{g_j},{g_k}}^ - } \right)}  \leqslant 1,\;\;{g_j} \in G, \label{eq-30}\\
& \chi _{{g_j},{g_k}}^ + ,\chi _{{g_j},{g_k}}^ -  \in \left\{ {0,1} \right\},\;\;{g_j},{g_k} \in G,\;\;j < k, \label{eq-31}\\
& {{\mathbf{u}^\text{INT}}^{\text{T}}}{{{\mathbf{V}}^{{\text{INT}}}}\left( {{a^*}} \right)} = 1, \label{eq-32}\\
& d \geqslant 0, \label{eq-33}\\
& {\mathbf{u}^\text{INT}} \geqslant {\mathbf{0}}, \label{eq-34}
\end{align}
where ${{\mathbf{S}}^{{\text{INT}}}} = \sum\limits_{k = 1}^q {\sum\limits_{a,b \in A_k^R} {\left( {{{\mathbf{V}}^{{\text{INT}}}}\left( a \right) - {{\mathbf{V}}^{{\text{INT}}}}\left( b \right)} \right){{\left( {{{\mathbf{V}}^{{\text{INT}}}}\left( a \right) - {{\mathbf{V}}^{{\text{INT}}}}\left( b \right)} \right)}^{\text{T}}}} }$, and ${\boldsymbol{\mu }}_k^{{\text{INT}}} = \frac{1}
{{\left| {A_k^R} \right|}}\sum\limits_{a \in A_k^R} {{{\mathbf{V}}^{{\text{INT}}}}\left( a \right)}$. $\chi _{{g_j},{g_k}}^ +$ and $\chi _{{g_j},{g_k}}^ -$ are binary variables such that $\chi _{{g_j},{g_k}}^ + = 1$ if positive interaction between criteria $g_j$ and $g_k$ exists, or $\chi _{{g_j},{g_k}}^ + = 0$, and $\chi _{{g_j},{g_k}}^ - = 1$ if negative interaction between criteria $g_j$ and $g_k$ exists, or otherwise $\chi _{{g_j},{g_k}}^ - = 0$. In constraint (\ref{eq-28}), $\chi _{{g_j},{g_k}}^ +$ and $\chi _{{g_j},{g_k}}^ -$ are used to identify whether interaction between criteria $g_j$ and $g_k$ exists, i.e., if there are some $\eta _{{g_j},{g_k}}^{ + ,s,t}$ or $\eta _{{g_j},{g_k}}^{ + ,s,t}$ greater than zero, then $\chi _{{g_j},{g_k}}^ + = 1$ or $\chi _{{g_j},{g_k}}^ - = 1$. Constraint (\ref{eq-29}) states that $syn_{{g_j},{g_k}}^ + \left( { \cdot , \cdot } \right)$ and $syn_{{g_j},{g_k}}^ - \left( { \cdot , \cdot } \right)$ are mutually exclusive, i.e., there exists only one type of interaction (i.e., positive or negative) between criteria $g_j$ and $g_k$. Constraint (\ref{eq-30}) ensures that any criterion interacts with at most one other. ${{\mathbf{V}}^{{\text{INT}}}}\left( {{a^*}} \right)$ is constructed according to a virtual ideal alternative ${{a^*}}$ such that ${g_j}\left( {{a^*}} \right) = {\beta _j}$, $j = 1,...,n$, and constraint (\ref{eq-32}) is used to bound ${U \left( \cdot \right)}$ to the interval [0,1]. 

Remark that the regularization term $\left\| {\mathbf{u}^\text{INT}} \right\|_2^2$ not only makes the derived marginal value functions as ``smooth'' as possible as in model (P2), but also smooths the variations of interaction parameters $\eta _{{g_j},{g_k}}^{ + ,s,t},\eta _{{g_j},{g_k}}^{ - ,s,t}$ over the adjacent grids constituted by criteria $g_j$ and $g_k$'s sub-intervals. Although model (P3) involves binary variables $\chi _{{g_j},{g_k}}^ +$ and $\chi _{{g_j},{g_k}}^ -$ for each pair of criteria $g_j$ and $g_k$, it is a convex quadratic optimization problem when the binary variables are fixed. Thus, it can be addressed in reasonable time using popular optimization packages if the number of criteria is not prohibitively large.

Model (P3) discovers the interactions among criteria from the given data and identifies the type of interaction (positive or negative) for any pair of interacting criteria in an automatic way. For some problems, it is reasonable to consider only one type of interaction for the interacting criteria (positive or negative). This can be implemented by adding the following constraint to Model (P3) in case only positive interaction is considered:
\begin{align*}
\chi _{{g_j},{g_k}}^ -  = 0,\;\;{g_j},{g_k} \in G,\;j < k,
\end{align*}
whereas the following one can be added when only negative interaction is admissible:
\begin{align*}
\chi _{{g_j},{g_k}}^ +  = 0,\;\;{g_j},{g_k} \in G,\;j < k.
\end{align*}

\subsection{Classification algorithms}
\begin{figure}[!htbp] 
	\centering
	\scalebox{1} {\includegraphics[width=0.7\textwidth]{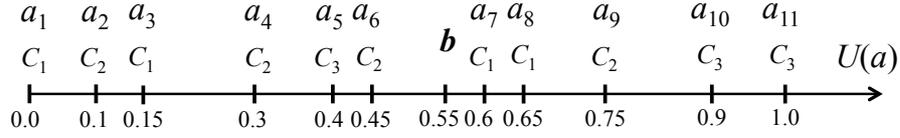}}
	\caption{\label{fig:assex1}Alternatives used to illustrate the use of four classification methods in Example 1.}
\end{figure}

\begin{figure}[!htbp] 
	\centering
	\scalebox{1} {\includegraphics[width=0.7\textwidth]{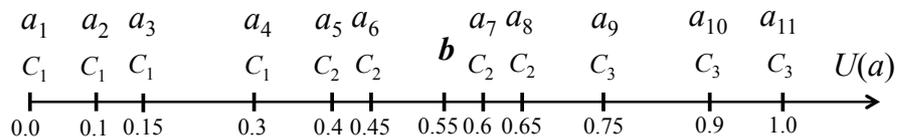}}
	\caption{\label{fig:assex2}Alternatives used to illustrate the use of four classification methods in Example 2.}
\end{figure}

\noindent Once the optimal value of $\textbf{u}$ or $\mathbf{u}^\text{INT}$ is obtained, we compute the comprehensive values for all reference ($a \in A^R$) and non-reference ($a \in A$) alternatives according to $U(a) = \textbf{u}^\text{T} \textbf{V}(a)$ or $U(a) = {{\mathbf{u}^\text{INT}}^{\text{T}}}{\mathbf{V}^\text{INT}}(a)$. To perform the classification of the latter ones, we will use four sorting methods denoted by $M_I$ -- $M_{IV}$. When determining the assignment for $a$, these methods work out the coefficients $M_r\left({a \to {Cl_k}} \right)$, $r \in \{I, II, III, IV\}$, indicating a~support given to $a \to {Cl_k}$ for $k=1,\ldots,q$, and select class $Cl_t$ for which $M_r\left({a \to {Cl_t}} \right)$ is maximal. We will explain the proposed approaches by referring to a pair of example sorting problems presented in Figures~\ref{fig:assex1} and~\ref{fig:assex2}. Each of these problems involves $11$ non-reference alternatives $a_{1,\ldots,11} \in A^R$ with a desired class specified in the respective figure as well as a single non-reference alternative $b \in A$ whose classification is yet to be determined.

The first classification method ($M_I$) was proposed with the aim of measuring the support given to a~hypothesis $a \to {Cl_k}$, $k=1,\ldots,q$, with the following consistency degree:
\begin{equation*}
M_I\left( {a \to {Cl_k}} \right) = \frac{{\left| {\left\{ {{a^*} \in \bigcup_{t=1,\ldots,k-1} A_t^R\; : \;U\left( a \right) > U\left( {{a^*}} \right)} \right\}} \right| + \left| {\left\{ {{a^*} \in \bigcup_{t=k+1,\ldots, q} A_{t}^R \; : \;U\left( a \right) < U\left( {{a^*}} \right)} \right\}} \right|}}
{{\left| \bigcup_{t=1,\ldots,q, \; t \ne k} {A_t^R} \right|}}.
\end{equation*}
The measure indicates a proportion of reference alternatives assigned to a~class either worse or better than $Cl_k$ that attain comprehensive values, respectively, lower or greater than $a$ according to the inferred value function. Clearly, the greater $M_I(a \rightarrow Cl_k)$ is, the lesser is the proportion of reference alternatives suggesting an assignment of $a$ to a class different than $Cl_k$, and hence the more justified is $a \to {Cl_k}$. For example, when considering an assignment of alternative $b$ presented in Figure~\ref{fig:assex1} to class $Cl_3$, $M_I\left({b \to {Cl_3}} \right) = 5/8$, because 5~($a_1, a_2, a_3, a_4, a_6$) out of $8$ reference alternatives assigned to class $Cl_1$ or $Cl_2$ (i.e., different than $Cl_3$) admit $b \to {Cl_3}$. Since $M_I\left({b \to {Cl_1}} \right) = 3/7$ and $M_I\left({b \to {Cl_2}} \right) = 4/7$, $Cl_3$ has the greatest support and $M_I$ would assign $b$ to $Cl_3$.

Method $M_I$ assumes that when confirming a given assignment each reference alternative has the same voting power, irrespective of its own consistency. In turn, method $M_{II}$ differentiates these powers by considering for each $a \in  A_k^R$ its consistency with the remaining assignment examples quantified with $M_I\left( {a \to {Cl_k}} \right)$. For example, when accounting for $a_4$ and $a_{10}$ from Figure~\ref{fig:assex1}, $M_I\left( {a_4\to {Cl_2}} \right) = 5/7$ and $M_I\left( {a_{10}\to {Cl_3}} \right) = 8/8$. Hence, we revise the coefficient used in method $M_{I}$ by assigning a~greater voting power to the reference alternatives which are more in line with other assignment examples:
\begin{equation*}
M_{II} \left( {a \to {Cl_k}} \right) = \frac{ \sum_{ {\left\{ {{a^*} \in \bigcup_{t=1,\ldots,k-1} A_t^R\; : \;U\left( a \right) > U\left( {{a^*}} \right)} \right\}} \bigcup {\left\{ {{a^*} \in \bigcup_{t=k+1,\ldots, q} A_{t}^R \; : \;U\left( a \right) < U\left( {{a^*}} \right)} \right\}}} M_I\left( {a^* \to {Cl_t}} \right) }
{{\left| \bigcup_{t=1,\ldots,q, \; t \ne k} {A_t^R} \right|}}.
\end{equation*}
For example, when considering an assignment of $b$ from Figure~\ref{fig:assex1} to $Cl_3$, $M_{II}\left({b \to {Cl_3}} \right) = (7/7 + 4/7 + 6/7 + 5/7 + 4/7)/8 = 26/56$, because 5 ($a_1, a_2, a_3, a_4, a_6$) out of $8$ reference alternatives assigned to $Cl_1$ or $Cl_2$ admit $b \to {Cl_3}$, but their consistency degrees do differ ($M_I\left( {a_1\to {Cl_1}} \right) = 7/7$, $M_I\left( {a_2\to {Cl_2}} \right) = 4/7$, $M_I\left( {a_3\to {Cl_1}} \right) = 6/7$, $M_I\left( {a_4\to {Cl_2}} \right) = 5/7$, $M_I\left( {a_6\to {Cl_2}} \right) = 4/7$). Since $M_{II}\left({b \to {Cl_1}} \right) = 20/49$ and $M_{II}\left({b \to {Cl_2}} \right) = 27/49$, $M_{II}$ would assign $b$ to $Cl_2$.

When computing the support given to $a \to {Cl_k}$, $M_I$ and $M_{II}$ consider the sets of reference alternatives assigned to classes different than $Cl_k$. On the contrary, method $M_{III}$ accounts for the reference alternatives contained in $A_k^R$, hence directly supporting $a \to {Cl_k}$. Moreover, it considers the consistencies of these reference alternatives according to their roles in view of validating $a \to {Cl_k}$. On one hand, for $a^* \in A_k^R$ such that $U(a^*) \le U(a)$, we analyse the subset of reference alternatives with comprehensive values not lesser than $U(a^*)$ and not greater than $U(a)$. They all have an impact on the worst possible class of $a$, hence influencing the strength of support given to $a \to {Cl_k}$ by $a^* \in A_k^R$. On the other hand, for $a^* \in A_k^R$ such that $U(a^*) > U(a)$, we analyse the subset of reference alternatives with comprehensive values not greater than $U(a^*)$ and greater than $U(a)$, as these have an impact on the best possible class of $a$. Overall, each $a^* \in A_k^R$ with either $U(a^*) \le U(a)$ or $U(a^*) > U(a)$ supports $a \to {Cl_k}$ with a~degree proportional to a~number of reference alternatives $a^{'} \in A_k^R$ with comprehensive values $U(a^*) \le U(a^{'}) \le U(a)$ or $U(a^*) \ge U(a^{'}) > U(a)$, respectively. For example, $a_5$~from Figure~\ref{fig:assex1} supports $b \to {Cl_3}$ with the strength of $1/2$, because out of $2$ reference alternatives $a^{'}$ with $U(a_5) \le U(a^{'}) \le U(b)$ (i.e., $a_5$ and $a_6$) only $a_5$ was assigned to $Cl_3$. In the same spirit, $a_{11}$ supports $b \to {Cl_3}$ with the strength of $2/5$, because out of $5$ reference alternatives $a^{'}$ with $U(a_{11}) \ge U(a^{'}) > U(b)$ (i.e., $a_7$, $a_8$, $a_9$, $a_{10}$, and $a_{11}$) only $a_{10}$ and $a_{11}$ were assigned to $Cl_3$. An overall support in favor of assignment $a \to {Cl_k}$, $k=1,\ldots,q$, can be quantified as an average support given by all $a^* \in A_k^R$ to $a \to {Cl_k}$, i.e.:
\begin{equation*}
M_{III} \left( {a \to {Cl_k}} \right) = \frac{ \sum_{ {\left\{ {{a^*} \in A_k^R\; : \;U\left( a \right) \ge U\left( {{a^*}} \right)} \right\}} } \frac{|A^R_{\lbrack a^*,a \rbrack} \bigcap A^R_k|}{|A^R_{\lbrack a^*,a \rbrack}|} + \sum_{ {\left\{ {{a^*} \in A_{k}^R \; : \;U\left( a \right) < U\left( {{a^*}} \right)} \right\}}} \frac{|A^R_{( a,a^* \rbrack} \bigcap A^R_k|}{|A^R_{(a,a^* \rbrack}|} }
{{\left| A_k^R \right|}},
\end{equation*}
where $A^R_{\lbrack a^*,a \rbrack} = \{ a^{'} \in A^R : U(a^*) \le U(a^{'}) \le U(a) \}$ and $A^R_{(a,a^* \rbrack} = \{  a^{'} \in A^R :  U(a) < U(a^{'}) \le U(a^*) \}$). When considering an assignment of $b$ from Figure~\ref{fig:assex1} to $Cl_3$, $M_{III}\left({b \to {Cl_3}} \right) = (1/2 + 1/4 + 2/5)/3 = 23/60$. As far as 3 reference alternatives assigned to $Cl_3$ ($a_5, a_{10}, a_{11}$) are concerned in view of $b \to Cl_3$, $a_5$ has a comprehensive value not greater than $U(b)$, but its consistency degree is $1/2$, because in $A^R_{\lbrack a_5, b \rbrack} = \{a_5, a_6\}$ only $1$ out of $2$ reference alternatives was assigned to $Cl_3$. Moreover, $a_{10}$ and $a_{11}$ have comprehensive values greater than $U(b)$ and their respective consistency degrees in terms of supporting $b \to Cl_3$ are $1/4$ ($A^R_{( b, a_{10} \rbrack} = \{a_7, a_8, a_9, a_{10}\}$) and $2/5$ ($A^R_{( b, a_{11} \rbrack} = \{a_7, a_8, a_9, a_{10}, a_{11}\}$). Since $M_{III}\left({b \to {Cl_1}} \right) = 31/48$ and $M_{III}\left({b \to {Cl_2}} \right) = 39/60$, $b$ is assigned to $Cl_2$ by $M_{III}$.

The last method ($M_{IV}$) analyses the support given to the assignment of $a \in A$ to different classes only by the reference alternatives which are the closest to $a$ in terms of their comprehensive values. Thus, we implement an idea originally postulated in the $K$-Nearest Neighbour ($K$-NN) algorithm, and additionally make the power of support given by each of $K$ closest reference alternatives equal to the reciprocal of an absolute value difference from $U(a)$. In this way, the reference alternatives with very similar scores to $a$ have a greater impact on its recommended assignment, i.e.:
\begin{equation*}
M_{IV}^K \left( {a \to {Cl_k}} \right) = \frac{ \sum_{a^* \in A^R_{a,K} \bigcap A^R_{k}} \frac{1}{|U(a) - U(a^*)|} }
{\sum_{a^* \in A^R_{a,K}} \frac{1}{|U(a) - U(a^*)|}},
\end{equation*}
where $A^R_{a,K}$ is a subset of $K$ reference alternatives with comprehensive values being the closest to $U(a)$. For example, when considering $b$ from Figure~\ref{fig:assex1}, $A^R_{b,K=3} = \{a_6, a_7, a_8\}$ with $a_6$ being assigned to $Cl_2$ and $a_7$ as well as $a_8$ being assigned to $Cl_1$. Thus, the supports given to the assignment of $b$ to different classes are as follows: $M_{IV}^{K=3} \left( {b \to {Cl_1}} \right) = (1/0.05 + 1/0.1)/(1/0.1 + 1/0.05 + 1/0.1) = 30/40$, $M_{IV}^{K=3} \left( {b \to {Cl_2}} \right) = (1/0.1)/(1/0.1 + 1/0.05 + 1/0.1) = 10/40$, $M_{IV}^{K=3} \left( {b \to {Cl_3}} \right) = 0/(1/0.1 + 1/0.05 + 1/0.1) = 0/40$, and $M_{IV}$ assigns $b$ to $Cl_1$. We can also use a~cross-validation to determine the optimal setting for $K$: for each possible value of $K$, classify all alternatives $a$ in the validation set by calculating $M_{IV}^K \left( {a \to {Cl_k}} \right)$, $k=1,...,q$, and then report the classification accuracy on the validation set. Value of $K$ that leads to the best classification performance can be chosen as the optimal setting for $K$. Note that $K$ should never be greater than the cardinality of the least numerous class so that the comparison of $M_{IV}^K \left( {a \to {Cl_k}} \right)$ across classes is performed fairly.

The assignments of reference alternatives $a_{1,\ldots,11} \in A^R$ used in Example 1 (see Figure~\ref{fig:assex1}) were intentionally selected so that to illustrate the differences in the sorting recommendation suggested for $b$ by various methods (for $M_I$ -- $Cl_3$, for $M_{II}$ and $M_{III}$ -- $Cl_2$, and for $M_{IV}$ -- $Cl_1$). On the contrary, when considering Example~2 (see Figure~\ref{fig:assex2}) with the consistency being preserved for all pairs of reference alternatives, the assignments suggested for $b$ by the four methods are the same. The detailed confirmation degrees given by $M_I$ -- $M_{IV}$ to different classes are provided in Table~\ref{tab:assex2}. Note that in case all assignment examples are perfectly reproduced, the scores provided by $M_I$ and $M_{II}$ are the same for all classes.

\begin{table}[!htbp]\caption{\label{tab-1}Confirmation degrees given by methods $M_I$ -- $M_{IV}$ to classes $Cl_1 - Cl_2$ when considering assignment of alternative $b$ from Example 2.}
	\centering 	\label{tab:assex2}
	\footnotesize
	\begin{tabular}{c c c c c}
		\hline
		Method & $Cl_1$ & $Cl_2$ & $Cl_3$ & Assignment  \\
		\hline
		$M_{I}$ & $5/7$ & $7/7$ & $6/8$ & $Cl_2$  \\
		$M_{II}$ & $5/7$ & $7/7$ & $6/8$ & $Cl_2$  \\
		$M_{III}$ & $126/240$ & $4/4$ & $43/90$ & $Cl_2$  \\
		$M_{IV}^{K=3}$ & $0/40$ & $40/40$ & $0/40$ & $Cl_2$ \\	
		\hline	
	\end{tabular}
\end{table}

\section{Experimental analysis}
\label{sec-experimental-analysis}

\subsection{Application to research unit evaluation}
\label{sec-experimental-analysis-1}
\noindent In this section, we apply the proposed approach to a real-world problem of parametric evaluation of research units. The Polish Ministry of Science and Higher Education carries out an evaluation of 993 units every 3 years and assigns each unit to one of four classes $C_1 - C_4$ (with $C_1$ and $C_4$ being, respectively, the least and the most preferred ones). All research units are divided into the following five subsets: 282 units from Humanities and Social Sciences (HS), 218 units from Life Sciences (NZ), 286 units from Exact Sciences and Engineering (SI), 99 units from Art and Artistic Creativity (TA), and 108 units judged as Inhomogeneous (NJN). They are evaluated in terms of four gain-type criteria: \emph{scientific activity ($g_1$)}, \emph{scientific potential ($g_2$)}, \emph{material effects of unit's activities ($g_3$)}, and \emph{remaining (non-material) effects of unit's activities ($g_4$)}. Even though all units are evaluated on the same family of criteria, the authorities apply different evaluation strategies for each subset of units to reflect the specificity of various fields of science. The data considered in this section is originally available at the website\footnote{http://www.bip.nauka.gov.pl/kategorie-naukowe-przyznane-jednostkom-naukowym/wyniki-kompleksowej-oceny-jakosci-dzialalnosci-naukowej-lub-badawczo-rozwojowej-jednostek-naukowych-2017.html}.

To validate the performance of proposed approach on this problem, we use five-fold cross-validation for the purpose of model selection and performance validation. Specifically, each subset of the dataset corresponding to a particular field of science (HS, NZ, SI, TA and NJN) is randomly partitioned into five folds of equal size. 
Then, four folds are used as the training data to construct a model, and the remaining one is used to test the performance of the constructed model. In the training phase, one of the four training folds serves as the validation data to determine the optimal setting for the two parameters $C_1$ and $C_2$ by examining the following values:\{${10^{ - 8}}$, $5 \times {10^{ - 8}}$, ${10^{ - 7}}$, $5 \times {10^{ - 7}}$,..., ${10^7}$, $5 \times {10^7}$, ${10^8}$, $5 \times {10^8}$\}. We repeat this procedure five times using different training and test sets. Finally, we average the results from each test and report a summary of the averaged performance.

The classification accuracies of the proposed approach parametrized with different numbers of sub-intervals on each criterion ($\gamma_j = 1,2,3$), preference models (${\Gamma}_{1}$: the variant neglecting interactions among criteria; ${\Gamma}_{2}$: the counterpart modelling the interaction effects in the product form (\ref{eq-22})-(\ref{eq-23}); ${\Gamma}_{3}$: the counterpart modelling the interaction effects in the minimum form (\ref{eq-22-1})-(\ref{eq-23-1})) and sorting methods ($M_{I}$, $M_{II}$, $M_{III}$ and $M_{IV}$) are summarized in Table \ref{table-0}. For the sake of simplicity, we suppose that all criteria have the same number of sub-intervals. For sorting method $M_{IV}$, we determine the optimal setting for $K$ using cross-validation by considering the following values: \{1,2,...,10\}. As can be observed from Table \ref{table-0}, by considering the interactions among criteria, the classification performance of the proposed approach improves as it is more flexible in fitting the training data. The variant of the proposed approach modelling ghe interaction effects in the product form (\ref{eq-22})-(\ref{eq-23}) achieves a better performance than its counterpart incorporating the minimum form (\ref{eq-22-1})-(\ref{eq-23-1}). The primary reason consists in that interaction terms in the product form (\ref{eq-22})-(\ref{eq-23}) are strictly increasing with respect to each component, thus having a higher ability in capturing the interactions among criteria than that in the minimum form (\ref{eq-22-1})-(\ref{eq-23-1}). As far as the comparison between different sorting methods ($M_{I}$, $M_{II}$, $M_{III}$ and $M_{IV}$) is concerned, it is evident that $M_{II}$ outperforms $M_{I}$, since $M_{II}$ differentiates the voting powers of reference alternatives by considering their consistency. Furthermore, the advantage of $M_{II}$, $M_{III}$ and $M_{IV}$ over their competitors is unclear. Finally, the performance of proposed approach is acceptable when coupled with different numbers of sub-intervals on each criterion, since we use the regularization terms to avoid the problem of over-fitting. However, the classification accuracy attained with $\gamma_j = 2$ is slightly higher than for the remaining values of $\gamma_j$.

\begin{table}[!htbp] \caption{\label{table-0}Classification accuracy of the proposed approach for the problem of parametric evaluation of Polish research units (${\Gamma}_{1}$: the variant neglecting the interactions among criteria; ${\Gamma}_{2}$: the counterpart modelling the interaction effects in the product form (\ref{eq-22})-(\ref{eq-23}); ${\Gamma}_{3}$: the counterpart modelling the interaction effects in the minimum form (\ref{eq-22-1})-(\ref{eq-23-1})).}
	\centering
	\scriptsize
	\begin{tabular}{cc ccc ccc ccc cc}
		\hline
		\multirow{2}[2]{*}{Subset} & \multirow{2}[2]{*}{Sorting
			method} & \multicolumn{3}{c}{${\gamma_{j}}=1$} &       & \multicolumn{3}{c}{${\gamma_{j}}=2$} &       & \multicolumn{3}{c}{${\gamma_{j}}=3$} \\
		\cmidrule{3-5}\cmidrule{7-9}\cmidrule{11-13}      &       & ${\Gamma}_{1}$ & ${\Gamma}_{2}$ & ${\Gamma}_{3}$ &       & ${\Gamma}_{1}$ & ${\Gamma}_{2}$ & ${\Gamma}_{3}$ &       & ${\Gamma}_{1}$ & ${\Gamma}_{2}$ & ${\Gamma}_{3}$ \\
		\hline
		\multirow{4}[0]{*}{HS} & $M_{I}$   & 0.8484  & 0.8587  & 0.8557  &       & 0.8987  & 0.9216  & 0.9159  &       & 0.7996  & 0.8640  & 0.8060  \\
		& $M_{II}$  & 0.8398  & 0.8870  & 0.8726  &       & 0.9556  & 0.9600  & 0.9586  &       & 0.8241  & 0.8839  & 0.8365  \\
		& $M_{III}$ & 0.8778  & 0.8665  & 0.8733  &       & 0.8961  & 0.9471  & 0.9406  &       & 0.8336  & 0.8804  & 0.8707  \\
		& $M_{IV}$  & 0.8638  & 0.8842  & 0.8768  &       & 0.8669  & 0.9334  & 0.9241  &       & 0.8192  & 0.8689  & 0.8435  \\
		\multicolumn{13}{l}{} \\
		\multirow{4}[0]{*}{NZ} & $M_{I}$   & 0.8992  & 0.8829  & 0.8937  &       & 0.8450  & 0.8904  & 0.8490  &       & 0.8677  & 0.8872  & 0.8858  \\
		& $M_{II}$  & 0.9136  & 0.9041  & 0.9073  &       & 0.8961  & 0.9295  & 0.9001  &       & 0.8420  & 0.9019  & 0.8917  \\
		& $M_{III}$ & 0.8694  & 0.8858  & 0.8729  &       & 0.8394  & 0.9011  & 0.8744  &       & 0.8817  & 0.8899  & 0.8829  \\
		& $M_{IV}$  & 0.8874  & 0.8975  & 0.8883  &       & 0.8938  & 0.9216  & 0.9074  &       & 0.8250  & 0.8906  & 0.8467  \\
		\multicolumn{13}{l}{} \\
		\multirow{4}[0]{*}{SI} & $M_{I}$   & 0.7703  & 0.8106  & 0.7976  &       & 0.8065  & 0.8324  & 0.8264  &       & 0.7830  & 0.8286  & 0.8024  \\
		& $M_{II}$  & 0.8059  & 0.8153  & 0.8078  &       & 0.8267  & 0.8512  & 0.8347  &       & 0.8100  & 0.8568  & 0.8489  \\
		& $M_{III}$ & 0.7636  & 0.8125  & 0.7827  &       & 0.8313  & 0.8463  & 0.8451  &       & 0.7986  & 0.8549  & 0.8176  \\
		& $M_{IV}$  & 0.7844  & 0.8129  & 0.7871  &       & 0.7876  & 0.8457  & 0.8255  &       & 0.8227  & 0.8526  & 0.8303  \\
		\multicolumn{13}{l}{} \\
		\multirow{4}[0]{*}{TA} & $M_{I}$   & 0.8435  & 0.8654  & 0.8456  &       & 0.8792  & 0.8891  & 0.8856  &       & 0.8394  & 0.8730  & 0.8579  \\
		& $M_{II}$  & 0.8463  & 0.8827  & 0.8567  &       & 0.8865  & 0.8921  & 0.8873  &       & 0.9141  & 0.9139  & 0.9141  \\
		& $M_{III}$ & 0.8544  & 0.8744  & 0.8589  &       & 0.9026  & 0.8907  & 0.8949  &       & 0.8952  & 0.8862  & 0.8933  \\
		& $M_{IV}$  & 0.8715  & 0.8812  & 0.8802  &       & 0.8559  & 0.8915  & 0.8637  &       & 0.9013  & 0.8886  & 0.8927  \\
		\multicolumn{13}{l}{} \\
		\multirow{4}[0]{*}{NJN} & $M_{I}$   & 0.7967  & 0.8218  & 0.8150  &       & 0.7872  & 0.8545  & 0.7912  &       & 0.8299  & 0.8228  & 0.8290  \\
		& $M_{II}$  & 0.8272  & 0.8543  & 0.8344  &       & 0.8749  & 0.8752  & 0.8749  &       & 0.7997  & 0.8529  & 0.8200  \\
		& $M_{III}$ & 0.8575  & 0.8537  & 0.8539  &       & 0.8539  & 0.8606  & 0.8599  &       & 0.8547  & 0.8489  & 0.8530  \\
		& $M_{IV}$  & 0.7913  & 0.8247  & 0.8010  &       & 0.8248  & 0.8562  & 0.8327  &       & 0.7883  & 0.8410  & 0.7901  \\
		\hline
	\end{tabular}
\end{table}

The proposed approach provides useful information about both the importance of individual criteria and the interactions between pairs of criteria. Since the accounted decision problem involves few criteria, we can analyse the constructed preference models in an intuitive way. For illustrative purpose, let us present the models constructed for the HS subset. The models generated by $\Gamma_1$, $\Gamma_2$ and $\Gamma_3$ with the optimal classification performance (denoted by $U^{\Gamma_1}\left( a \right)$, $U^{\Gamma_2}\left( a \right)$ and $U^{\Gamma_3}\left( a \right)$, respectively) are presented in the following. One can observe that, when neglecting the interactions among criteria, the weights of the four criteria are $w_{g_1}=\Delta u_{{g_{1}}}^1 + \Delta u_{{g_{1}}}^2=0.4198$, $w_{g_2}=\Delta u_{{g_{2}}}^1 + \Delta u_{{g_{2}}}^2=0.2151$, $w_{g_3}=\Delta u_{{g_{3}}}^1 + \Delta u_{{g_{3}}}^2=0.1957$ and $w_{g_4}=\Delta u_{{g_{4}}}^1 + \Delta u_{{g_{4}}}^2=0.1694$, respectively:
\begin{equation*}
\begin{gathered}
U^{\Gamma_1}\left( a \right) = 0.1724 \cdot v_{{g_1}}^1\left( a \right) + 0.2474 \cdot v_{{g_1}}^2\left( a \right) + 0.0806 \cdot v_{{g_2}}^1\left( a \right) + 0.1345 \cdot v_{{g_2}}^2\left( a \right) \hfill \\
\;\;\;\;\;\; + 0.0612 \cdot v_{{g_3}}^1\left( a \right) + 0.1345 \cdot v_{{g_3}}^2\left( a \right) + 0.0557 \cdot v_{{g_4}}^1\left( a \right) + 0.1137 \cdot v_{{g_4}}^2\left( a \right). \hfill \\ 
\end{gathered}
\end{equation*}
\noindent In case of modelling criteria interactions in the product form, the weights are $w_{g_1}=0.6656$, $w_{g_2}=0.3891$, $w_{g_3}=0.1462$ and $w_{g_4}=0.3177$, respectively, and there are negative interactions for the pairs $\{g_1, g_3\}$ and $\{g_2,g_4\}$:

\begin{equation*}
\begin{gathered}
U^{\Gamma_2}\left( a \right) = 0.2850 \cdot v_{{g_1}}^1\left( a \right) + 0.3806 \cdot v_{{g_1}}^2\left( a \right) + 0.1526 \cdot v_{{g_2}}^1\left( a \right) + 0.2365 \cdot v_{{g_2}}^2\left( a \right) \hfill \\
\;\;\;\;\;\; + 0.0506 \cdot v_{{g_3}}^1\left( a \right) + 0.0956 \cdot v_{{g_3}}^2\left( a \right) + 0.0482 \cdot v_{{g_4}}^1\left( a \right) + 0.2695 \cdot v_{{g_4}}^2\left( a \right) \hfill \\
\;\;\;\;\;\; - 0.1298 \cdot v_{{g_1}}^1\left( a \right) \cdot v_{{g_3}}^1\left( a \right) - 0.0 \cdot v_{{g_1}}^1\left( a \right) \cdot v_{{g_3}}^2\left( a \right) \hfill \\
\;\;\;\;\;\; - 0.0 \cdot v_{{g_1}}^2\left( a \right) \cdot v_{{g_3}}^1\left( a \right) - 0.0 \cdot v_{{g_1}}^2\left( a \right) \cdot v_{{g_3}}^2\left( a \right) \hfill \\
\;\;\;\;\;\; - 0.1526 \cdot v_{{g_2}}^1\left( a \right) \cdot v_{{g_4}}^1\left( a \right) - 0.0 \cdot v_{{g_2}}^1\left( a \right) \cdot v_{{g_4}}^2\left( a \right) \hfill \\
\;\;\;\;\;\; - 0.2365 \cdot v_{{g_2}}^2\left( a \right) \cdot v_{{g_4}}^1\left( a \right) - 0.0 \cdot v_{{g_2}}^2\left( a \right) \cdot v_{{g_4}}^2\left( a \right). \hfill \\ 
\end{gathered}
\end{equation*}

\noindent When it comes to the case of criteria interactions modelled in the minimum form, we derive the criteria weights as $w_{g_1}=0.4220$, $w_{g_2}=0.2164$, $w_{g_3}=0.1979$ and $w_{g_4}=0.1707$, respectively, and there exist positive interaction between $g_1$ and $g_4$ and negative interaction between $g_2$ and $g_3$:

\begin{equation*}
\begin{gathered}
U^{\Gamma_3}\left( a \right) = 0.1735 \cdot v_{{g_1}}^1\left( a \right) + 0.2485 \cdot v_{{g_1}}^2\left( a \right) + 0.0807 \cdot v_{{g_2}}^1\left( a \right) + 0.1357 \cdot v_{{g_2}}^2\left( a \right) \hfill \\
\;\;\;\;\;\; + 0.0623 \cdot v_{{g_3}}^1\left( a \right) + 0.1356 \cdot v_{{g_3}}^2\left( a \right) + 0.0568 \cdot v_{{g_4}}^1\left( a \right) + 0.1139 \cdot v_{{g_4}}^2\left( a \right) \hfill \\
\;\;\;\;\;\; + 0.2092 \cdot v_{{g_1}}^1\left( a \right) \cdot v_{{g_4}}^1\left( a \right) + 0.0 \cdot v_{{g_1}}^1\left( a \right) \cdot v_{{g_4}}^2\left( a \right) \hfill \\
\;\;\;\;\;\; + 0.0 \cdot v_{{g_1}}^2\left( a \right) \cdot v_{{g_4}}^1\left( a \right) + 0.0 \cdot v_{{g_1}}^2\left( a \right) \cdot v_{{g_4}}^2\left( a \right) \hfill \\
\;\;\;\;\;\; - 0.0395 \cdot \min \left\{ {v_{{g_2}}^1\left( a \right),v_{{g_3}}^1\left( a \right)} \right\} - 0.0411 \cdot \min \left\{ {v_{{g_2}}^1\left( a \right),v_{{g_3}}^2\left( a \right)} \right\} \hfill \\
\;\;\;\;\;\; - 0.0683 \cdot \min \left\{ {v_{{g_2}}^2\left( a \right),v_{{g_3}}^1\left( a \right)} \right\} - 0.0674 \cdot \min \left\{ {v_{{g_2}}^2\left( a \right),v_{{g_3}}^2\left( a \right)} \right\}. \hfill \\ 
\end{gathered}
\end{equation*}

\subsection{Experimental evaluation on several monotone learning datasets}
\label{sec-experimental-analysis-2}

\noindent In this section, we report the results of an extensive experimental study performed to validate the practical performance of the proposed approach on several public datasets. The goal of this study is two-fold. First, we compare the two variants of the proposed approach which either incorporate the interactions between criteria or neglect them with the UTADIS method and the Choquet integral-based sorting model in terms of a~set of predictive metrics. Second, we investigate the information about the relative importance of criteria and the interactions among criteria discovered by the proposed approach.

The datasets used in the experiments were collected from the UCI repository\footnote{http://archive.ics.uci.edu/ml/} and the WEKA machine learning datasets\footnote{http://www.cs.waikato.ac.nz/ml/weka/datasets.html}. For these datasets, after removing incomplete instances and descriptive attributes (e.g., name or gender), the assumption of monotonicity can be made on the remaining variables, so that the input and output variables can be deemed as criteria and decision classes, respectively. A summary of the information about these datasets is reported in Table~\ref{table-1}. The accounted datasets can be accessed at~\url{https://cs.uni-paderborn.de/?id=63916}.

\begin{table}[!htbp] \caption{\label{table-1}Datasets used in the experimental study and their properties.}
	\centering
	\scriptsize
	\begin{tabular}{l c c c l}
		\hline
		Dataset & \#Alternatives & \#Criteria & \#Classes & Distribution of classes\\
		\hline
		Den Bosch (DBS) & 120 & 8 & 2 & 60/60 \\
		CPU & 209 & 6 & 4 & 50/53/53/53 \\
		Breast Cancer (BCC) & 278 & 7 & 2 & 196/82 \\
		Auto MPG (MPG) & 392 & 7 & 3 & 107/154/131 \\
		Employee Selection (ESL) & 488 & 4 & 5 & 52/100/116/135/85 \\
		Mammographic (MMG) & 830 & 5 & 2 & 427/403 \\
		Employee Rejection/Acceptance (ERA) & 1000 & 4 & 3 & 415/330/255 \\
		Lecturers Evaluation (LEV) & 1000 & 4 & 4 & 93/280/403/224 \\
		Car Evaluation (CEV) & 1728 & 6 & 3 & 1210/384/134 \\
		\hline
	\end{tabular}
\end{table}

\noindent We use the same experimental setting as in Section \ref{sec-experimental-analysis-1} for model selection and performance validation. The evaluation metrics for quantifying the predictive performance of classification methods include~\emph{accuracy}, \emph{precision}, \emph{recall}, and \emph{F-measure} for detailed definitions of these metrics). The adopted version of the UTADIS method estimates an additive value function model by minimizing the total misclassification errors for all reference alternatives. Then, it uses a post optimality analysis to explore other or near optimal solutions, and finally works out a preference model by averaging all found solutions. The Choquet integral-based sorting model -- being presented in Appendix C -- is implemented by replacing the preference model in the proposed approach with the Choquet integral. UTADIS, the Choquet integral-based sorting model and the proposed approach are implemented in Java and the optimization models are solved using the Cplex solver\footnote{https://www.ibm.com/analytics/cplex-optimizer}. We repeat the cross-validation procedure 100 times, and then report the average of the results.

\begin{table}[!htbp] \caption{\label{table-2}Classification performance of UTADIS, Choquet integral-based sorting model, and the proposed approach (Type I: the proposed approach neglecting the interactions among criteria; Type II: the proposed approach considering the interactions among criteria).}
	\centering
	\scriptsize
	\begin{tabular}{crcccc}
		\hline
		 &  &  & Choquet integral- & \multicolumn{2}{c}{Proposed approach}  \\
		Dataset & Metrics & UTADIS & based approach & Type I & Type II \\
		\hline
		\multirow{4}[0]{*}{DBS} & Accuracy & 0.8247  & 0.8534  & 0.8362  & 0.8746  \\
		& Precision Avg. & 0.8258  & 0.8527  & 0.8401  & 0.8724  \\
		& Recall Avg. & 0.8249  & 0.8513  & 0.8377  & 0.8712  \\
		& F-measure Avg. & 0.8248  & 0.8523  & 0.8381  & 0.8724  \\
		\multicolumn{6}{l}{} \\
		\multirow{4}[0]{*}{CPU} & Accuracy & 0.9248  & 0.9273  & 0.9196  & 0.9388  \\
		& Precision Avg. & 0.9231  & 0.9278  & 0.9169  & 0.9426  \\
		& Recall Avg. & 0.9234  & 0.9252  & 0.9250  & 0.9359  \\
		& F-measure Avg. & 0.9232  & 0.9261  & 0.9214  & 0.9381  \\
		\multicolumn{6}{l}{} \\
		\multirow{4}[0]{*}{BCC} & Accuracy & 0.7106  & 0.7294  & 0.7466  & 0.7552  \\
		& Precision Avg. & 0.6514  & 0.6924  & 0.7296  & 0.7447  \\
		& Recall Avg. & 0.6469  & 0.7096  & 0.7739  & 0.7953  \\
		& F-measure Avg. & 0.6489  & 0.6861  & 0.7282  & 0.7385  \\
		\multicolumn{6}{l}{} \\
		\multirow{4}[0]{*}{MPG} & Accuracy & 0.8426  & 0.8837  & 0.8859  & 0.8939  \\
		& Precision Avg. & 0.8457  & 0.8819  & 0.8818  & 0.8874  \\
		& Recall Avg. & 0.8511  & 0.8853  & 0.8905  & 0.8889  \\
		& F-measure Avg. & 0.8410  & 0.8800  & 0.8841  & 0.8880  \\
		\multicolumn{6}{l}{} \\
		\multirow{4}[0]{*}{ESL} & Accuracy & 0.8607  & 0.8941  & 0.8641  & 0.9042  \\
		& Precision Avg. & 0.8481  & 0.8536  & 0.8565  & 0.8858  \\
		& Recall Avg. & 0.8610  & 0.8460  & 0.8684  & 0.8967  \\
		& F-measure Avg. & 0.8531  & 0.8423  & 0.8612  & 0.8901  \\
		\multicolumn{6}{l}{} \\
		\multirow{4}[0]{*}{MMG} & Accuracy & 0.8284  & 0.8627  & 0.8579  & 0.8751  \\
		& Precision Avg. & 0.8295  & 0.8736  & 0.8588  & 0.8900  \\
		& Recall Avg. & 0.8307  & 0.8660  & 0.8592  & 0.8782  \\
		& F-measure Avg. & 0.8288  & 0.8623  & 0.8585  & 0.8723  \\
		\multicolumn{6}{l}{} \\
		\multirow{4}[0]{*}{ERA} & Accuracy & 0.8106  & 0.8336  & 0.8501  & 0.8830  \\
		& Precision Avg. & 0.8030  & 0.8329  & 0.8428  & 0.8897  \\
		& Recall Avg. & 0.8098  & 0.8347  & 0.8496  & 0.8842  \\
		& F-measure Avg. & 0.8038  & 0.8325  & 0.8453  & 0.8863  \\
		\multicolumn{6}{l}{} \\
		\multirow{4}[0]{*}{LEV} & Accuracy & 0.8302  & 0.9010  & 0.8960  & 0.9173  \\
		& Precision Avg. & 0.8274  & 0.8924  & 0.8891  & 0.9080  \\
		& Recall Avg. & 0.8362  & 0.9105  & 0.9019  & 0.9278  \\
		& F-measure Avg. & 0.8281  & 0.9040  & 0.8935  & 0.9208  \\
		\multicolumn{6}{l}{} \\
		\multirow{4}[0]{*}{CEV} & Accuracy & 0.8912  & 0.9219  & 0.9243  & 0.9423  \\
		& Precision Avg. & 0.8798  & 0.9238  & 0.9147  & 0.9485  \\
		& Recall Avg. & 0.8954  & 0.9174  & 0.9298  & 0.9352  \\
		& F-measure Avg. & 0.8845  & 0.9187  & 0.9201  & 0.9410  \\
		\hline
	\end{tabular}
\end{table}

The comprehensive experimental results of the different variants of the proposed approach are reported in Appendix D. In the main paper, we report the results of the variant that achieves the highest classification accuracy (see Table \ref{table-2}) for the following analysis. To save space, we average the precisions, recalls and F-measures for all classes and report only the means of these measures. As can be observed in Table \ref{table-2}, the proposed approach compares favourably with UTADIS and the Choquet integral-based model for most datasets. In particular, the variant of the proposed approach accounting for the interactions among criteria outperforms its counterpart without considering such interactions. 

Furthermore, we perform the one-tailed paired $t$-test to examine a statistical significance of the observed differences. For each $t$-test, the null-hypothesis is ${H_0}:\;{\mu _{{\text{method}}\;{\text{1}}}} \leqslant {\mu _{{\text{method}}\;2}}$ and the alternative hypothesis is ${H_1}:\;{\mu _{{\text{method}}\;{\text{1}}}} > {\mu _{{\text{method}}\;2}}$, where ${\mu _{{\text{method}}\;{\text{1}}}}$ and ${\mu _{{\text{method}}\;2}}$ denote the means of a~measure for method 1 and method 2, respectively. Note that $p$-values less than a reasonable significance level 0.05 indicate that the null-hypothesis should be rejected and the alternative hypothesis is acceptable, and thus method 1 outperforms method 2 in terms of the underlying metric. The results of $t$-test are reported in Table~\ref{table-3}. They confirm both the~competitive advantage of our approach over UTADIS the Choquet integral-based model for most datasets as well as the superiority of the variant accounting for the interactions among criteria over the method's counterpart neglecting such interactions.

\begin{table}[!htbp] \caption{\label{table-3}Results of one-tailed paired $t$-test for comparing performances of different methods -- ($^*$) denotes significance at 5\% level (Comparison I: the proposed approach neglecting the interactions among criteria vs. UTADIS; Comparison II: the proposed approach considering the interactions among criteria vs. UTADIS; Comparison III: the proposed approach considering the interactions among criteria vs. Choquet integral-based approach; Comparison IV: the proposed approach considering the interactions among criteria vs. that neglecting the interactions among criteria.).}
	\centering
	\scriptsize
	\begin{tabular}{c r c c c c }
		\hline
		Dataset & Metrics & \multicolumn{1}{l}{Comparison I} & \multicolumn{1}{l}{Comparison II} & \multicolumn{1}{l}{Comparison III} & \multicolumn{1}{l}{Comparison IV} \\
		\hline
		\multirow{4}[0]{*}{DBS} & Accuracy & 0.0001*  & 0.0000*  & 0.0001*  & 0.0000*  \\
		& Precision Avg. & 0.0001*  & 0.0000*  & 0.0000*  & 0.0000*  \\
		& Recall Avg. & 0.0000*  & 0.0000*  & 0.0000*  & 0.0000*  \\
		& F-measure Avg. & 0.0000*  & 0.0000*  & 0.0000*  & 0.0000*  \\
		\multicolumn{6}{c}{} \\
		\multirow{4}[0]{*}{CPU} & Accuracy & 0.0647  & 0.0003*  & 0.0175*  & 0.0000*  \\
		& Precision Avg. & 0.0972  & 0.0000*  & 0.0000*  & 0.0000*  \\
		& Recall Avg. & 0.0738  & 0.0128*  & 0.0033*  & 0.0011*  \\
		& F-measure Avg. & 0.1619  & 0.0000*  & 0.0000*  & 0.0000*  \\
		\multicolumn{6}{c}{} \\
		\multirow{4}[0]{*}{BCC} & Accuracy & 0.0000*  & 0.0000*  & 0.0000*  & 0.0001*  \\
		& Precision Avg. & 0.0000*  & 0.0000*  & 0.0000*  & 0.0000*  \\
		& Recall Avg. & 0.0000*  & 0.0000*  & 0.0000*  & 0.0000*  \\
		& F-measure Avg. & 0.0000*  & 0.0000*  & 0.0000*  & 0.0001*  \\
		\multicolumn{6}{c}{} \\
		\multirow{4}[0]{*}{MPG} & Accuracy & 0.0000*  & 0.0000*  & 0.0000* & 0.0145*  \\
		& Precision Avg. & 0.0000*  & 0.0000*  & 0.0672  & 0.0526  \\
		& Recall Avg. & 0.0000*  & 0.0000*  & 0.0049*  & 0.1103  \\
		& F-measure Avg. & 0.0000*  & 0.0000*  & 0.0005*  & 0.0783  \\
		\multicolumn{6}{c}{} \\
		\multirow{4}[0]{*}{ESL} & Accuracy & 0.1633  & 0.0000*  & 0.0110*  & 0.0000*  \\
		& Precision Avg. & 0.0099*  & 0.0000*  & 0.0000*  & 0.0000*  \\
		& Recall Avg. & 0.0021*  & 0.0000*  & 0.0000*  & 0.0000*  \\
		& F-measure Avg. & 0.0000*  & 0.0000*  & 0.0000*  & 0.0000*  \\
		\multicolumn{6}{c}{} \\
		\multirow{4}[0]{*}{MMG} & Accuracy & 0.0000*  & 0.0000*  & 0.0000*  & 0.0000*  \\
		& Precision Avg. & 0.0000*  & 0.0000*  & 0.0001*  & 0.0000*  \\
		& Recall Avg. & 0.0000*  & 0.0000*  & 0.0000*  & 0.0000*  \\
		& F-measure Avg. & 0.0000*  & 0.0000*  & 0.0010*  & 0.0003*  \\
		\multicolumn{6}{c}{} \\
		\multirow{4}[0]{*}{ERA} & Accuracy & 0.0000*  & 0.0000*  & 0.0000*  & 0.0000*  \\
		& Precision Avg. & 0.0000*  & 0.0000*  & 0.0000*  & 0.0000*  \\
		& Recall Avg. & 0.0000*  & 0.0000*  & 0.0000*  & 0.0000*  \\
		& F-measure Avg. & 0.0000*  & 0.0000*  & 0.0000*  & 0.0000*  \\
		\multicolumn{6}{c}{} \\
		\multirow{4}[0]{*}{LEV} & Accuracy & 0.0000*  & 0.0000*  & 0.0000*  & 0.0000*  \\
		& Precision Avg. & 0.0000*  & 0.0000*  & 0.0002*  & 0.0000* \\
		& Recall Avg. & 0.0000*  & 0.0000*  & 0.0000*  & 0.0000*  \\
		& F-measure Avg. & 0.0000*  & 0.0000*  & 0.0000*  & 0.0000*  \\
		\multicolumn{6}{c}{} \\
		\multirow{4}[0]{*}{CEV} & Accuracy & 0.0000*  & 0.0000*  & 0.0000*  & 0.0000*  \\
		& Precision Avg. & 0.0000*  & 0.0000*  & 0.0000*  & 0.0000*  \\
		& Recall Avg. & 0.0000*  & 0.0000*  & 0.0000*  & 0.0072*  \\
		& F-measure Avg. & 0.0000*  & 0.0000*  & 0.0000*  & 0.0000*  \\
		\hline
	\end{tabular}
\end{table}

\noindent The advantage of the proposed approach over UTADIS may be due to two reasons. First, the fitting ability of UTADIS improves with the increase in the number of sub-intervals on criteria, but it may encounter the over-fitting problem, leading to a poor predictive accuracy on test data. In the proposed approach, the incorporation of the regularization term alleviates the over-fitting problem and enhances the generalization ability to test data. Second, when compared with UTADIS and the counterpart neglecting the interactions among criteria, the variant of the proposed approach accounting for the interactions is more flexible and has the ability for capturing non-linear dependencies among criteria. When it comes to the Choquet integral-based model, although it has a higher flexibility in modelling the interactions among criteria, it requires the performance on all criteria to be defined on the same scale, which is a great limitation of this preference model. In the experimental analysis, we have brought all performances to the same scale through a scaling method (see Appendix C), which assumes that the marginal value functions on all criteria have the same form. This deteriorates the fitting ability of the preference model on several datasets. 

We present the relative importances of all criteria and the interaction coefficients in Table~\ref{table-4}, which can be obtained by calculating the marginal values on each criterion and the bonus and penalty coefficients of interacting criteria for a virtual alternative $a^*$ with the optimal performances on each criterion (i.e., ${g_j}\left( {{a^*}} \right) = {\beta _j}$, $j=1,...,n$). They correspond to the preference models worked out by the proposed approach achieving the highest test accuracy on each dataset. For example, for the DBS dataset, when neglecting the interactions, we can order the eight criteria according to their relative importance as follows: $g_4$ $>$ $g_2$ $>$ $g_8$ $>$ $g_5$ $>$ $g_1$ $>$ $g_3$ $>$ $g_6$ $>$ $g_7$. When considering the interactions, one can observe that the order of criteria according to their relative importances remains the same and additionally there exist positive interactions for pairs $\{g_1,g_5\}$ and $\{g_3,g_7\}$ and negative interactions for pairs $\{g_2,g_8\}$ and $\{g_4,g_6\}$.

\begin{table}[!htbp] \caption{\label{table-4}Criteria weights and interaction coefficients discovered by the proposed approach (Type I: results obtained from proposed approach neglecting the interactions among criteria; Type II: results obtained from proposed approach considering the interactions among criteria; $w_{(\cdot)}$ indicates a weight of a particular criterion which is derived from marginal value of virtual alternative $a^*$ on this criterion; $\Phi_{(\cdot,\cdot)}$ indicates the interaction intensity between a pair of interacting criteria which is obtained from bonus or penalty value of virtual alternative $a^*$ on the two criteria).}
	\centering
	\scriptsize
	\begin{tabular}{c c l}
		\hline
		Dataset & Type  & Criteria weights and interaction coefficients \\
		\hline
		\multirow{2}[0]{*}{DBS} & Type I & \tabincell{l}{$w_{g_1}$=0.1044, $w_{g_2}$=0.1884, $w_{g_3}$=0.0772, $w_{g_4}$=0.2751, \\ $w_{g_5}$=0.1523, $w_{g_6}$=0.0258, $w_{g_7}$=0.0158, $w_{g_8}$=0.1610 \\ \hspace*{\fill}} \\
		& Type II & \tabincell{l}{$w_{g_1}$=0.0860, $w_{g_2}$=0.2405, $w_{g_3}$=0.0625, $w_{g_4}$=0.2885, \\ $w_{g_5}$=0.1032, $w_{g_6}$=0.0295, $w_{g_7}$=0.0187, $w_{g_8}$=0.1874, \\ $\Phi_{g_1,g_5}$=+0.0226, $\Phi_{g_2,g_8}$=--0.0418, $\Phi_{g_3,g_7}$=+0.0258, \\ $\Phi_{g_4,g_6}$=--0.0228} \\
		\multicolumn{3}{c}{} \\
		\multirow{2}[0]{*}{CPU} & Type I & \tabincell{l}{$w_{g_1}$=0.1915, $w_{g_2}$=0.1330, $w_{g_3}$=0.3285, $w_{g_4}$=0.1643, \\ $w_{g_5}$=0.0641, $w_{g_6}$=0.1186 \\ \hspace*{\fill}} \\
		& Type II & \tabincell{l}{$w_{g_1}$=0.1746, $w_{g_2}$=0.1519, $w_{g_3}$=0.3635, $w_{g_4}$=0.1487, \\ $w_{g_5}$=0.0409, $w_{g_6}$=0.1036, \\ $\Phi_{g_1,g_5}$=+0.0365, $\Phi_{g_2,g_3}$=--0.0480, $\Phi_{g_4,g_6}$=+0.0284} \\
		\multicolumn{3}{c}{} \\
		\multirow{2}[0]{*}{BCC} & Type I & \tabincell{l}{$w_{g_1}$=0.0309, $w_{g_2}$=0.1819, $w_{g_3}$=0.1997, $w_{g_4}$=0.1495, \\$w_{g_5}$=0.2988, $w_{g_6}$=0.0159, $w_{g_7}$=0.1231 \\ \hspace*{\fill}} \\
		& Type II & \tabincell{l}{$w_{g_1}$=0.0219, $w_{g_2}$=0.2004, $w_{g_3}$=0.2124, $w_{g_4}$=0.1423, \\ $w_{g_5}$=0.2799, $w_{g_6}$=0.0126, $w_{g_7}$=0.1432, \\ $\Phi_{g_1,g_5}$=+0.0107, $\Phi_{g_2,g_7}$=--0.0306, $\Phi_{g_3,g_6}$=+0.0073} \\
		\multicolumn{3}{c}{} \\
		\multirow{2}[0]{*}{MPG} & Type I & \tabincell{l}{$w_{g_1}$=0.1727, $w_{g_2}$=0.1278, $w_{g_3}$=0.0806, $w_{g_4}$=0.1088, \\ $w_{g_5}$=0.2545, $w_{g_6}$=0.0372, $w_{g_7}$=0.2186 \\ \hspace*{\fill}} \\
		& Type II & \tabincell{l}{$w_{g_1}$=0.1875, $w_{g_2}$=0.1099, $w_{g_3}$=0.0927, $w_{g_4}$=0.0820, \\ $w_{g_5}$=0.2284, $w_{g_6}$=0.0329, $w_{g_7}$=0.1952, \\ $\Phi_{g_1,g_3}$=--0.0215, $\Phi_{g_2,g_7}$=0.0508, $\Phi_{g_4,g_5}$=+0.0421} \\
		\multicolumn{3}{c}{} \\
		\multirow{2}[0]{*}{ESL} & Type I & \tabincell{l}{$w_{g_1}$=0.2201, $w_{g_2}$=0.2223, $w_{g_3}$=0.2904, $w_{g_4}$=0.2672 \\ \hspace*{\fill}} \\
		& Type II & \tabincell{l}{$w_{g_1}$=0.1969, $w_{g_2}$=0.2184, $w_{g_3}$=0.2648, $w_{g_4}$=0.2761, \\$\Phi_{g_1,g_3}$=+0.0855, $\Phi_{g_2,g_4}$=--0.0418} \\
		\multicolumn{3}{c}{} \\
		\multirow{2}[0]{*}{MMG} & Type I & \tabincell{l}{$w_{g_1}$=0.1369, $w_{g_2}$=0.3665, $w_{g_3}$=0.1923, $w_{g_4}$=0.2166, \\ $w_{g_5}$=0.0878 \\ \hspace*{\fill}} \\
		& Type II & \tabincell{l}{$w_{g_1}$=0.1430, $w_{g_2}$=0.3258, $w_{g_3}$=0.1745, $w_{g_4}$=0.2321, \\$w_{g_5}$=0.1041,  $\Phi_{g_1,g_4}$=--0.0877,$\Phi_{g_2,g_3}$=+0.1082} \\
		\multicolumn{3}{c}{} \\
		\multirow{2}[0]{*}{ERA} & Type I & \tabincell{l}{$w_{g_1}$=0.3123, $w_{g_2}$=0.4381, $w_{g_3}$=0.1560, $w_{g_4}$=0.0936 \\ \hspace*{\fill}} \\
		& Type II & \tabincell{l}{$w_{g_1}$=0.2928, $w_{g_2}$=0.3401, $w_{g_3}$=0.1661, $w_{g_4}$=0.1130, \\$\Phi_{g_1,g_2}$=+0.1254, $\Phi_{g_3,g_4}$=--0.0376} \\
		\multicolumn{3}{c}{} \\
		\multirow{2}[0]{*}{LEV} & Type I & \tabincell{l}{$w_{g_1}$=0.3197, $w_{g_2}$=0.4361, $w_{g_3}$=0.0509, $w_{g_4}$=0.1933 \\ \hspace*{\fill}} \\
		& Type II & \tabincell{l}{$w_{g_1}$=0.3288, $w_{g_2}$=0.4895, $w_{g_3}$=0.0632, $w_{g_4}$=0.2161, \\$\Phi_{g_1,g_2}$=--0.1008, $\Phi_{g_3,g_4}$=+0.0033} \\
		\multicolumn{3}{c}{} \\
		\multirow{2}[0]{*}{CEV} & Type I & \tabincell{l}{$w_{g_1}$=0.1624, $w_{g_2}$=0.1439, $w_{g_3}$=0.0411, $w_{g_4}$=0.2593, \\$w_{g_5}$=0.0854, $w_{g_6}$=0.3077 \\ \hspace*{\fill}} \\
		& Type II & \tabincell{l}{$w_{g_1}$=0.1407, $w_{g_2}$=0.1206, $w_{g_3}$=0.0805, $w_{g_4}$=0.2010, \\$w_{g_5}$=0.1192, $w_{g_6}$=0.2380, $\Phi_{g_1,g_2}$=+0.0467, \\ $\Phi_{g_3,g_5}$=--0.0328, $\Phi_{g_4,g_6}$=+0.0860} \\
		\hline
	\end{tabular}
\end{table}

To investigate the computational efforts needed by the proposed approach for addressing datasets of various sizes, we report its runtime in Table~\ref{table-5}. One can observe the results in a two-fold way. First, when neglecting the interactions between criteria, the proposed approach addresses all sorting problems in less than 25 seconds and there is no significant difference among the computation time of the proposed approach for dealing with different datasets although they involve various numbers of alternatives, criteria and classes. This is because, in model (P2), the reference alternatives in each class $C{l_k}$, $k=1,...,q$, are aggregated into the vector ${\boldsymbol{\mu }}_k$, and then the computation time for a specific problem is not related to the number of available reference alternatives. Moreover, since the numbers of criteria and classes for these datasets are small (less than 10), the convex quadratic programming model in (P2) can be easily solved in a~short time. Second, when considering the interactions among criteria, the runtime of the proposed approach increases sharply with the number of criteria. For the datasets involving four criteria (ESL, ERA, and LEV), the runtime is less than one minute, whereas for problems involving eight criteria (DBS), the computation time already exceeds 1000 seconds. The reason behind this phenomenon is that model (P3) involves a combinatorial optimization problem to find the pairs of interacting criteria and such a problem becomes more difficult to solved with greater number of criteria. In conclusion, the proposed approach is suitable for dealing with data-intensive tasks that contain a large number of alternatives, whereas the number of criteria -- in case of accounting for the potential interactions -- highly affects the computation time of the proposed approach.

\begin{table}[!htbp] \caption{\label{table-5}Computation time of the proposed approach for addressing different datasets (Type I: variant neglecting the interactions among criteria; Type II: counterpart considering the interactions among criteria).}
	\centering
	\scriptsize
	\begin{tabular}{llllll}
		\hline
		Dataset & Type  & Mean  & Std.  & Min.  & Max. \\
		\hline
		\multirow{2}[0]{*}{DBS} & Type I & 19.7550  & 1.2227  & 17.8809  & 21.5419  \\
		& Type II & 1012.6324  & 58.3988  & 910.6261  & 1114.1336  \\
		\multirow{2}[0]{*}{CPU} & Type I & 19.9319  & 1.2453  & 17.6707  & 22.1471  \\
		& Type II & 193.1181  & 10.4591  & 174.4472  & 210.1466  \\
		\multirow{2}[0]{*}{BCC} & Type I & 19.7922  & 1.0719  & 18.0529  & 21.6751  \\
		& Type II & 493.2548  & 29.3294  & 444.3109  & 540.7687  \\
		\multirow{2}[0]{*}{MPG} & Type I & 20.0035  & 1.8410  & 17.9479  & 21.9236  \\
		& Type II & 499.1771  & 20.5614  & 462.1352  & 540.1671  \\
		\multirow{2}[0]{*}{ESL} & Type I & 19.9830  & 1.2722  & 17.7951  & 22.1560  \\
		& Type II & 48.9712  & 2.4282  & 44.6182  & 53.5375  \\
		\multirow{2}[0]{*}{MMG} & Type I & 19.6550  & 1.1620  & 17.7482  & 21.7930  \\
		& Type II & 108.8784  & 5.9832  & 97.9610  & 119.0889  \\
		\multirow{2}[0]{*}{ERA} & Type I & 20.0823  & 1.1856  & 18.1521  & 22.1972  \\
		& Type II & 49.8116  & 3.2252  & 44.2085  & 55.1615  \\
		\multirow{2}[0]{*}{LEV} & Type I & 20.1161  & 0.9851  & 18.2588  & 21.6508  \\
		& Type II & 48.5154  & 2.7686  & 44.3528  & 53.4136  \\
		\multirow{2}[0]{*}{CEV} & Type I & 20.3521  & 1.1549  & 18.1207  & 22.1505  \\
		& Type II & 190.1866  & 12.9837  & 167.0662  & 211.7082  \\
		\hline
	\end{tabular}
\end{table}

\section{Conclusions}
\label{sec-conclusions}

\noindent In this paper, we considered multiple criteria sorting problems with potentially interacting criteria and proposed a new preference learning approach for constructing a preference model from a~given set of decision examples. We perceive the introduced method as a novel data-driven decision support tool, because the preference discovery is performed automatically without participation of the DM. As such, it connects the fields of Multiple Criteria Decision Aiding and Machine Learning. On the one hand, the proposed approach starts with an additive piecewise-linear value function as the preference model under the preferential independence hypothesis, and then enriches it with two types of components for modelling the positive and negative interactions between criteria. Therefore, the employed preference models have the advantage of comprehensibility and can be easily understood and accepted by the DM. On the other hand, the proposed approach utilizes methodological advances in Machine Learning to enhance the predictive ability of the constructed preference model and the computational efficiency of the preference learning procedure. Specifically, to improve the generalization ability of the constructed model on new decision instances, it uses regularization terms to avoid the problem of over-fitting. Therefore, it is robust with respect to the way of dividing the performance scale of each criterion into sub-intervals. In addition, the incorporated optimization models are formulated in such a~way that the complexity is irrelevant from the number of training samples, thus being capable of addressing data-intensive tasks.

The practical example of research unit evaluation illustrates the applicability of the proposed approach and intuitiveness of the arrived results. Furthermore, the experimental outcomes on several public datasets demonstrate its advantage over UTADIS and the Choquet integral-based sorting method in terms of both predictive performance and interpretability. Such a competitive advantage derives from accounting for the interactions between criteria without transforming all performances to the same scale, making the proposed approach suitable for capturing complex interdependencies in data. Moreover, the joint use of piecewise-linear marginal value functions and regularization terms improves the fitting ability and avoids the problem of over-fitting, thus making the generalization performance on new alternatives more advantageous.

Actually, our approach can be deemed as a general framework for addressing data-driven multiple criteria sorting problems. Within such a framework, we provide a set of components for implementing a preference learning procedure, including the variants accounting or neglecting the interactions among criteria, different types of expressions for quantifying the positive and negative interactions among criteria and a few sorting methods for classifying non-reference alternatives. One can select various components to equip different variants of the proposed approach which prove to be more advantageous for different learning tasks. However, one needs to be aware that a prerequisite of our approach is the monotonicity assumption on the input variables so that they can be regarded as criteria. Therefore, the proposed approach aims to learn a predictive model that guarantees monotonicity and considers interactions of input variables simultaneously. When such an assumption is not valid, one can consider non-monotonicity in the input variables as an alternative way for representing the complex patterns in the data.

Future research can investigate the following directions. We can extend the proposed approach to consider both interacting and non-monotonic criteria simultaneously. Moreover, other types of preference models can be extended to capture interactions between criteria and then studied in the preference learning framework for data-driven decision support. We will also develop computationally efficient algorithms for dealing with problems involving greater number of criteria. Finally, more real-world applications are needed to validate the performance of the proposed approach.

\section*{References}
\bibliographystyle{apalike}
\bibliography{mybibfile}

\section*{Appendix A. Proof of Proposition 1}
\noindent
\begin{description}
	\item[Proof.] Let $d_0^*$ and $d_1^*$ denote the optimal values of the objective functions of models (P0) and (P1), respectively. Suppose that $d_0^* \ne d_1^*$. Then, two possible cases would occur: (a): $d_0^* > d_1^*$, or (b): $d_0^* < d_1^*$. In case (a), let ${U^*}\left(  \cdot  \right)$ and ${d^*}\left( {a,b} \right)$ be, respectively, the value function and the difference between any pair of reference alternatives $(a,b) \in A_{s + 1}^R \times A_s^R$, $s = 1,...,q - 1$, at the optimum of model (P0). Thus, according to constraint (\ref{eq-5}), we have that ${U^*}\left( a \right) - {U^*}\left( b \right) = {d^*}\left( {a,b} \right),\;\;a \in A_{s + 1}^R,\;b \in A_s^R,\;s = 1,...,q - 1$. By taking the transformation (\ref{eq-7})-(\ref{eq-11}), we will derive that ${{\mathbf{u}}^*}^{\text{T}}{{\boldsymbol{\mu }}_{s + 1}} - {{\mathbf{u}}^*}^{\text{T}}{{\boldsymbol{\mu }}_s} = \frac{1}{{\left| {A_{s + 1}^R} \right|\left| {A_s^R} \right|}}\sum\nolimits_{a \in A_{s + 1}^R,\;b \in A_s^R} {{d^*}\left( {a,b} \right)} ,\;s = 1,...,q - 1$, where ${{\mathbf{u}}^*}$ is the value of ${\mathbf{u}}$ at the optimum. According to constraint (\ref{eq-6}), we have $d_0^* = \mathop {\min }\limits_{s = 1,...,q - 1} \frac{1}
	{{\left| {A_{s + 1}^R} \right|\left| {A_s^R} \right|}}\sum\nolimits_{a \in A_{s + 1}^R,\;b \in A_s^R} {{d^*}\left( {a,b} \right)} $. In this way, we find a value of ${\mathbf{u}}$ (i.e., ${{\mathbf{u}}^*}$) so that the minimum value of ${{\mathbf{u}}^{\text{T}}}{{\boldsymbol{\mu }}_{s + 1}} - {{\mathbf{u}}^{\text{T}}}{{\boldsymbol{\mu }}_s}$, $s = 1,...,q - 1$, is equal to $d_0^*$. As $d_0^* > d_1^*$, it contradicts that the optimal value of the objective function of model (P1) is $d_1^*$. Therefore, the assumption $d_0^* > d_1^*$ cannot hold. On the other hand, in case (b), let ${{\mathbf{u}}^*}$ be the value of ${\mathbf{u}}$ at the optimum of model (P1). Then, we use ${U^*}\left(  \cdot  \right)$ and ${d^*}\left( {a,b} \right)$	to denote the value function and the difference between any pair of reference alternatives $(a,b) \in A_{s + 1}^R \times A_s^R$, $s = 1,...,q - 1$, determined by ${{\mathbf{u}}^*}$. That is to say, ${U^*}\left( a \right) = {{\mathbf{u}}^*}^{\text{T}}{\mathbf{V}}\left( a \right)$, $a \in {A^R}$, and ${U^*}\left( a \right) - {U^*}\left( b \right) = {d^*}\left( {a,b} \right),\;\;a \in A_{s + 1}^R,\;b \in A_s^R,\;s = 1,...,q - 1$. Then, for $s = 1,...,q - 1$, we have ${{\mathbf{u}}^*}^{\text{T}}{{\boldsymbol{\mu }}_{s + 1}} - {{\mathbf{u}}^*}^{\text{T}}{{\boldsymbol{\mu }}_s}$ $ = {{\mathbf{u}}^*}^{\text{T}}\left( {\frac{1}
		{{\left| {A_{s + 1}^R} \right|}}\sum\nolimits_{a \in A_{s + 1}^R} {{\mathbf{V}}\left( a \right)} } \right) - {{\mathbf{u}}^*}^{\text{T}}\left( {\frac{1}
		{{\left| {A_s^R} \right|}}\sum\nolimits_{b \in A_s^R} {{\mathbf{V}}\left( b \right)} } \right)$ $ = \frac{1}
	{{\left| {A_{s + 1}^R} \right|}}\sum\nolimits_{a \in A_{s + 1}^R} {{U^*}\left( a \right)}  - \frac{1}
	{{\left| {A_s^R} \right|}}\sum\nolimits_{b \in A_s^R} {{U^*}\left( b \right)} $ $ = \frac{1}
	{{\left| {A_{s + 1}^R} \right|\left| {A_s^R} \right|}}\left\{ {\left| {A_s^R} \right|\sum\nolimits_{a \in A_{s + 1}^R} {{U^*}\left( a \right)}  - \left| {A_{s + 1}^R} \right|\sum\nolimits_{b \in A_s^R} {{U^*}\left( b \right)} } \right\}$ $ = \frac{1}
	{{\left| {A_{s + 1}^R} \right|\left| {A_s^R} \right|}}\sum\nolimits_{a \in A_{s + 1}^R,\;b \in A_s^R} {\left\{ {{U^*}\left( a \right) - {U^*}\left( b \right)} \right\}} $ $ = \frac{1}
	{{\left| {A_{s + 1}^R} \right|\left| {A_s^R} \right|}}\sum\nolimits_{a \in A_{s + 1}^R,\;b \in A_s^R} {{d^*}\left( {a,b} \right)} $. According to constraint (\ref{eq-13}), there must be some ${s^*} \in \left\{ {1,...,q - 1} \right\}$ such that ${{\mathbf{u}}^*}^{\text{T}}{{\boldsymbol{\mu }}_{{s^*} + 1}} - {{\mathbf{u}}^*}^{\text{T}}{{\boldsymbol{\mu }}_{{s^*}}}$$= \frac{1}
	{{\left| {A_{{s^*} + 1}^R} \right|\left| {A_{{s^*}}^R} \right|}}\sum\nolimits_{a \in A_{{s^*} + 1}^R,\;b \in A_{{s^*}}^R} {{d^*}\left( {a,b} \right)}  = d_1^*$, while for other $s \ne {s^*}$, we have ${{\mathbf{u}}^*}^{\text{T}}{{\boldsymbol{\mu }}_{s + 1}} - {{\mathbf{u}}^*}^{\text{T}}{{\boldsymbol{\mu }}_s} = \frac{1}
	{{\left| {A_{s + 1}^R} \right|\left| {A_s^R} \right|}}\sum\nolimits_{a \in A_{s + 1}^R,\;b \in A_s^R} {{d^*}\left( {a,b} \right)}  \geqslant d_1^*$. In this way, we find value function ${U^*}\left(  \cdot  \right)$ and the difference ${d^*}\left( {a,b} \right)$ between any pair of reference alternatives $(a,b) \in A_{s + 1}^R \times A_s^R$, $s = 1,...,q - 1$, such that the optimal value of the objective function of model (P0) is equal to $d_1^*$. As $d_1^* > d_0^*$, it contradicts that the optimal value of the objective function of model (P0) is $d_0^*$. Therefore, the assumption $d_0^* < d_1^*$ cannot hold. To sum up, the assumption $d_0^* \ne d_1^*$ cannot hold and it must be that $d_0^* = d_1^*$. \qed
\end{description}

\section*{Appendix B. Proof of Proposition 2}
\noindent
\begin{description}
	\item[Proof.] We only present the proof for the case of the bonus and penalty components being defined as (\ref{eq-22}) and (\ref{eq-23}), and another case of (\ref{eq-22-1}) and (\ref{eq-23-1}) can be analysed analogously.
		
	On the one hand, since $x_j^s > x_j^{s - 1}$, $s = 1,...,{\gamma _j}$, according to monotonicity (b), for any $\left\{ {{g_j},{g_k}} \right\} \in Syn$, we have:
	\begin{equation*}
	\begin{gathered}
	\left[ {{u_{{g_j}}}\left( {x_j^s} \right) + {u_{{g_k}}}\left( {x_k^t} \right) + \left( {syn_{{g_j},{g_k}}^ + \left( {x_j^s,x_k^t} \right) - syn_{{g_j},{g_k}}^ - \left( {x_j^s,x_k^t} \right)} \right)} \right] \hfill \\
	\;\;\;\;\;\; - \left[ {{u_{{g_j}}}\left( {x_j^{s - 1}} \right) + {u_{{g_k}}}\left( {x_k^t} \right) + \left( {syn_{{g_j},{g_k}}^ + \left( {x_j^{s - 1},x_k^t} \right) - syn_{{g_j},{g_k}}^ - \left( {x_j^{s - 1},x_k^t} \right)} \right)} \right] \geqslant 0, \hfill \\ 
	\end{gathered}
	\end{equation*}
	for $s = 1,...,{\gamma _j}$, $t = 1,...,{\gamma _k}$. Then, according the definitions of ${u_{{g_j}}}\left(  \cdot  \right)$, $syn_{{g_j},{g_k}}^ + \left( { \cdot , \cdot } \right)$, and $syn_{{g_j},{g_k}}^ - \left( { \cdot , \cdot } \right)$, the above inequality can be further transformed to:
	\begin{equation*}
	\begin{gathered}
	\left[ {{u_{{g_j}}}\left( {x_j^s} \right) + {u_{{g_k}}}\left( {x_k^t} \right) + \left( {syn_{{g_j},{g_k}}^ + \left( {x_j^s,x_k^t} \right) - syn_{{g_j},{g_k}}^ - \left( {x_j^s,x_k^t} \right)} \right)} \right] \hfill \\
	\;\;\;\;\;\; - \left[ {{u_{{g_j}}}\left( {x_j^{s - 1}} \right) + {u_{{g_k}}}\left( {x_k^t} \right) + \left( {syn_{{g_j},{g_k}}^ + \left( {x_j^{s - 1},x_k^t} \right) - syn_{{g_j},{g_k}}^ - \left( {x_j^{s - 1},x_k^t} \right)} \right)} \right] \hfill \\
	= \left[ {\sum\limits_{p = 1}^s {\Delta u_{{g_j}}^p}  + \sum\limits_{q = 1}^t {\Delta u_{{g_k}}^q}  + \left( {\sum\limits_{p = 1}^s {\sum\limits_{q = 1}^t {\eta _{{g_j},{g_k}}^{ + ,p,q}} }  - \sum\limits_{p = 1}^s {\sum\limits_{q = 1}^t {\eta _{{g_j},{g_k}}^{ - ,p,q}} } } \right)} \right] \hfill \\
	\;\;\;\;\;\; - \left[ {\sum\limits_{p = 1}^{s - 1} {\Delta u_{{g_j}}^p}  + \sum\limits_{q = 1}^t {\Delta u_{{g_k}}^q}  + \left( {\sum\limits_{p = 1}^{s - 1} {\sum\limits_{q = 1}^t {\eta _{{g_j},{g_k}}^{ + ,p,q}} }  - \sum\limits_{p = 1}^{s - 1} {\sum\limits_{q = 1}^t {\eta _{{g_j},{g_k}}^{ - ,p,q}} } } \right)} \right] \hfill \\
	= \Delta u_{{g_j}}^s + \sum\limits_{q = 1}^t {\left( {\eta _{{g_j},{g_k}}^{ + ,s,q} - \eta _{{g_j},{g_k}}^{ - ,s,q}} \right)}  \geqslant 0, \hfill \\ 
	\end{gathered}
	\end{equation*}
	for $s = 1,...,{\gamma _j}$, $t = 1,...,{\gamma _k}$.
	
	On the other hand, if $\Delta u_{{g_j}}^s + \sum\limits_{q = 1}^t {\left( {\eta _{{g_j},{g_k}}^{ + ,s,q} - \eta _{{g_j},{g_k}}^{ - ,s,q}} \right)}  \geqslant 0,\;\;s = 1,...,{\gamma _j},\;t = 1,...,{\gamma _k}$, $\forall \left\{ {{g_j},{g_k}} \right\} \in Syn$, then, for any pair of alternatives $a,b$ such that ${g_j}\left( a \right) \geqslant {g_j}\left( b \right)$ and ${g_k}\left( a \right) \geqslant {g_k}\left( b \right)$, $\forall \left\{ {{g_j},{g_k}} \right\} \in Syn$, let us suppose that ${g_j}\left( a \right) \in \left[ {x_j^{p - 1},x_j^p} \right]$, ${g_k}(a) \in \left[ {x_k^{q - 1},x_k^q} \right]$, ${g_j}\left( b \right) \in \left[ {x_j^{p' - 1},x_j^{p'}} \right]$ and ${g_k}(b) \in \left[ {x_k^{q' - 1},x_k^{q'}} \right]$. Obviously, $p \geqslant p'$ and $q \geqslant q'$. There are four possible cases: (a) $p = p',\;q = q'$, (b) $p > p',\;q = q'$, (c) $p = p',\;q > q'$, and (d) $p > p',\;q > q'$. In case (a), $\forall \left\{ {{g_j},{g_k}} \right\} \in Syn$, we can formulate
	\begin{equation*}
	\begin{gathered}
	\left[ {{u_{{g_j}}}\left( {{g_j}\left( a \right)} \right) + {u_{{g_k}}}\left( {{g_k}\left( a \right)} \right) + \left( {syn_{{g_j},{g_k}}^ + \left( {{g_j}\left( a \right),{g_k}\left( a \right)} \right) - syn_{{g_j},{g_k}}^ - \left( {{g_j}\left( a \right),{g_k}\left( a \right)} \right)} \right)} \right] \hfill \\
	\;\;\;\;\;\; - \left[ {{u_{{g_j}}}\left( {{g_j}\left( b \right)} \right) + {u_{{g_k}}}\left( {{g_k}\left( b \right)} \right) + \left( {syn_{{g_j},{g_k}}^ + \left( {{g_j}\left( b \right),{g_k}\left( b \right)} \right) - syn_{{g_j},{g_k}}^ - \left( {{g_j}\left( b \right),{g_k}\left( b \right)} \right)} \right)} \right] \hfill \\
	= \left[ {\left( {\sum\limits_{s = 1}^{p - 1} {\Delta u_{{g_j}}^s} {\text{ + }}\Delta u_{{g_j}}^pv_{{g_j}}^p\left( a \right)} \right) + \left( {\sum\limits_{t = 1}^{q - 1} {\Delta u_{{g_k}}^s} {\text{ + }}\Delta u_{{g_k}}^qv_{{g_k}}^q\left( a \right)} \right)} \right. + \left( {\sum\limits_{s = 1}^{p - 1} {\sum\limits_{t = 1}^{q - 1} {\left( {\eta _{{g_j},{g_k}}^{ + ,s,t} - \eta _{{g_j},{g_k}}^{ - ,s,t}} \right)} } } \right. \hfill \\
	\;\;\;\;\;\; + \left. {\left. {\sum\limits_{t = 1}^{q - 1} {\left( {\eta _{{g_j},{g_k}}^{ + ,p,t} - \eta _{{g_j},{g_k}}^{ - ,p,t}} \right)v_{{g_j}}^p\left( a \right)}  + \sum\limits_{s = 1}^{p - 1} {\left( {\eta _{{g_j},{g_k}}^{ + ,s,q} - \eta _{{g_j},{g_k}}^{ - ,s,q}} \right)} v_{{g_k}}^q\left( a \right) + \left( {\eta _{{g_j},{g_k}}^{ + ,p,q} - \eta _{{g_j},{g_k}}^{ - ,p,q}} \right)v_{{g_j}}^p\left( a \right)v_{{g_k}}^q\left( a \right)} \right)} \right] \hfill \\
	\;\;\;\;\;\; - \left[ {\left( {\sum\limits_{s = 1}^{p - 1} {\Delta u_{{g_j}}^s} {\text{ + }}\Delta u_{{g_j}}^pv_{{g_j}}^p\left( b \right)} \right) + \left( {\sum\limits_{t = 1}^{q - 1} {\Delta u_{{g_k}}^s} {\text{ + }}\Delta u_{{g_k}}^qv_{{g_k}}^q\left( b \right)} \right)} \right. + \left( {\sum\limits_{s = 1}^{p - 1} {\sum\limits_{t = 1}^{q - 1} {\left( {\eta _{{g_j},{g_k}}^{ + ,s,t} - \eta _{{g_j},{g_k}}^{ - ,s,t}} \right)} } } \right. \hfill \\
	\;\;\;\;\;\; + \left. {\left. {\sum\limits_{t = 1}^{q - 1} {\left( {\eta _{{g_j},{g_k}}^{ + ,p,t} - \eta _{{g_j},{g_k}}^{ - ,p,t}} \right)v_{{g_j}}^p\left( b \right)}  + \sum\limits_{s = 1}^{p - 1} {\left( {\eta _{{g_j},{g_k}}^{ + ,s,q} - \eta _{{g_j},{g_k}}^{ - ,s,q}} \right)} v_{{g_k}}^q\left( b \right) + \left( {\eta _{{g_j},{g_k}}^{ + ,p,q} - \eta _{{g_j},{g_k}}^{ - ,p,q}} \right)v_{{g_j}}^p\left( b \right)v_{{g_k}}^q\left( b \right)} \right)} \right] \hfill \\
	= \Delta u_{{g_j}}^p\left( {v_{{g_j}}^p\left( a \right) - v_{{g_j}}^p\left( b \right)} \right) + \Delta u_{{g_k}}^q\left( {v_{{g_k}}^q\left( a \right) - v_{{g_k}}^q\left( b \right)} \right) + \sum\limits_{t = 1}^{q - 1} {\left( {\eta _{{g_j},{g_k}}^{ + ,p,t} - \eta _{{g_j},{g_k}}^{ - ,p,t}} \right)} \left( {v_{{g_j}}^p\left( a \right) - v_{{g_j}}^p\left( b \right)} \right) \hfill \\
	\;\;\;\;\;\; + \sum\limits_{s = 1}^{p - 1} {\left( {\eta _{{g_j},{g_k}}^{ + ,s,q} - \eta _{{g_j},{g_k}}^{ - ,s,q}} \right)} \left( {v_{{g_k}}^q\left( a \right) - v_{{g_k}}^q\left( b \right)} \right) + \left( {\eta _{{g_j},{g_k}}^{ + ,p,q} - \eta _{{g_j},{g_k}}^{ - ,p,q}} \right)\left( {v_{{g_j}}^p\left( a \right)v_{{g_k}}^q\left( a \right) - v_{{g_j}}^p\left( b \right)v_{{g_k}}^q\left( b \right)} \right) \hfill \\
	= \left( {v_{{g_j}}^p\left( a \right) - v_{{g_j}}^p\left( b \right)} \right)\left( {\Delta u_{{g_j}}^p + \sum\limits_{t = 1}^{q - 1} {\left( {\eta _{{g_j},{g_k}}^{ + ,p,t} - \eta _{{g_j},{g_k}}^{ - ,p,t}} \right)} } \right) + \left( {v_{{g_k}}^q\left( a \right) - v_{{g_k}}^q\left( b \right)} \right)\left( {\Delta u_{{g_k}}^q + \sum\limits_{s = 1}^{p - 1} {\left( {\eta _{{g_j},{g_k}}^{ + ,s,q} - \eta _{{g_j},{g_k}}^{ - ,s,q}} \right)} } \right) \hfill \\
	\;\;\;\;\;\; + \left( {\eta _{{g_j},{g_k}}^{ + ,p,q} - \eta _{{g_j},{g_k}}^{ - ,p,q}} \right)\left( {v_{{g_j}}^p\left( a \right)v_{{g_k}}^q\left( a \right) - v_{{g_j}}^p\left( b \right)v_{{g_k}}^q\left( b \right)} \right). \hfill \\ 
	\end{gathered} 
	\end{equation*}
	Note that $v_{{g_j}}^p\left( a \right) - v_{{g_j}}^p\left( b \right) \geqslant 0$, $v_{{g_k}}^q\left( a \right) - v_{{g_k}}^q\left( b \right) \geqslant 0$, and $v_{{g_j}}^p\left( a \right)v_{{g_k}}^q\left( a \right) - v_{{g_j}}^p\left( b \right)v_{{g_k}}^q\left( b \right) \geqslant 0$, because ${g_j}\left( a \right) \geqslant {g_j}\left( b \right)$ and ${g_k}\left( a \right) \geqslant {g_k}\left( b \right)$. Moreover, $\Delta u_{{g_j}}^p + \sum\limits_{t = 1}^{q - 1} {\left( {\eta _{{g_j},{g_k}}^{ + ,p,t} - \eta _{{g_j},{g_k}}^{ - ,p,t}} \right)}  \geqslant 0$ and $\Delta u_{{g_k}}^q + \sum\limits_{s = 1}^{p - 1} {\left( {\eta _{{g_j},{g_k}}^{ + ,s,q} - \eta _{{g_j},{g_k}}^{ - ,s,q}} \right)}  \geqslant 0$. Then, let us consider the following quadratic multi-variate function:
	\begin{equation*}
	\begin{gathered}
	f\left( {v_{{g_j}}^p\left( a \right),v_{{g_j}}^p\left( b \right),v_{{g_k}}^q\left( a \right),v_{{g_k}}^q\left( b \right)} \right) = \left( {v_{{g_j}}^p\left( a \right) - v_{{g_j}}^p\left( b \right)} \right)\left( {\Delta u_{{g_j}}^p + \sum\limits_{t = 1}^{q - 1} {\left( {\eta _{{g_j},{g_k}}^{ + ,p,t} - \eta _{{g_j},{g_k}}^{ - ,p,t}} \right)} } \right) \hfill \\
	\;\;\;\;\;\; + \left( {v_{{g_k}}^q\left( a \right) - v_{{g_k}}^q\left( b \right)} \right)\left( {\Delta u_{{g_k}}^q + \sum\limits_{s = 1}^{p - 1} {\left( {\eta _{{g_j},{g_k}}^{ + ,s,q} - \eta _{{g_j},{g_k}}^{ - ,s,q}} \right)} } \right) \hfill \\
	\;\;\;\;\;\; + \left( {\eta _{{g_j},{g_k}}^{ + ,p,q} - \eta _{{g_j},{g_k}}^{ - ,p,q}} \right)\left( {v_{{g_j}}^p\left( a \right)v_{{g_k}}^q\left( a \right) - v_{{g_j}}^p\left( b \right)v_{{g_k}}^q\left( b \right)} \right), \hfill \\ 
	\end{gathered}
	\end{equation*}
	whose Hessian matrix is given as:
	\begin{equation*}
	{\mathbf{H}} = \left[ {\begin{array}{*{20}{c}}
		0 & 0 & {\left( {\eta _{{g_j},{g_k}}^{ + ,p,q} - \eta _{{g_j},{g_k}}^{ - ,p,q}} \right)} & 0  \\
		0 & 0 & 0 & { - \left( {\eta _{{g_j},{g_k}}^{ + ,p,q} - \eta _{{g_j},{g_k}}^{ - ,p,q}} \right)}  \\
		{\left( {\eta _{{g_j},{g_k}}^{ + ,p,q} - \eta _{{g_j},{g_k}}^{ - ,p,q}} \right)} & 0 & 0 & 0  \\
		0 & { - \left( {\eta _{{g_j},{g_k}}^{ + ,p,q} - \eta _{{g_j},{g_k}}^{ - ,p,q}} \right)} & 0 & 0  \\			
		\end{array} } \right].
	\end{equation*}
	${\mathbf{H}}$ is positive semi-definite, and thus $f\left( {v_{{g_j}}^p\left( a \right),v_{{g_j}}^p\left( b \right),v_{{g_k}}^q\left( a \right),v_{{g_k}}^q\left( b \right)} \right)$ is convex. Then, the minimum value of $f\left( {v_{{g_j}}^p\left( a \right),v_{{g_j}}^p\left( b \right),v_{{g_k}}^q\left( a \right),v_{{g_k}}^q\left( b \right)} \right)$ can be found in Table \ref{tab-app-1}. It can be observed that all the potential optimal values reported in Table \ref{tab-app-1} are greater than zero. Thus, $\forall \left\{ {{g_j},{g_k}} \right\} \in Syn$, we have
	\begin{equation*}
	\begin{gathered}
	\left[ {{u_{{g_j}}}\left( {{g_j}\left( a \right)} \right) + {u_{{g_k}}}\left( {{g_k}\left( a \right)} \right) + \left( {syn_{{g_j},{g_k}}^ + \left( {{g_j}\left( a \right),{g_k}\left( a \right)} \right) - syn_{{g_j},{g_k}}^ - \left( {{g_j}\left( a \right),{g_k}\left( a \right)} \right)} \right)} \right] \hfill \\
	\;\;\;\;\;\; \geqslant  \left[ {{u_{{g_j}}}\left( {{g_j}\left( b \right)} \right) + {u_{{g_k}}}\left( {{g_k}\left( b \right)} \right) + \left( {syn_{{g_j},{g_k}}^ + \left( {{g_j}\left( b \right),{g_k}\left( b \right)} \right) - syn_{{g_j},{g_k}}^ - \left( {{g_j}\left( b \right),{g_k}\left( b \right)} \right)} \right)} \right]. \hfill \\
	\end{gathered}
	\end{equation*}
	In case (b), utilizing the results from case (a), we can prove:
	\begin{equation*}
	\begin{gathered}
	\left[ {{u_{{g_j}}}\left( {{g_j}\left( a \right)} \right) + {u_{{g_k}}}\left( {{g_k}\left( a \right)} \right) + \left( {syn_{{g_j},{g_k}}^ + \left( {{g_j}\left( a \right),{g_k}\left( a \right)} \right) - syn_{{g_j},{g_k}}^ - \left( {{g_j}\left( a \right),{g_k}\left( a \right)} \right)} \right)} \right] \hfill \\
	\;\;\;\;\;\; \geqslant \left[ {{u_{{g_j}}}\left( {x_j^{p - 1}} \right) + {u_{{g_k}}}\left( {{g_k}\left( a \right)} \right) + \left( {syn_{{g_j},{g_k}}^ + \left( {x_j^{p - 1},{g_k}\left( a \right)} \right) - syn_{{g_j},{g_k}}^ - \left( {x_j^{p - 1},{g_k}\left( a \right)} \right)} \right)} \right] \hfill \\
	\;\;\;\;\;\; \geqslant \left[ {{u_{{g_j}}}\left( {x_j^{p'}} \right) + {u_{{g_k}}}\left( {{g_k}\left( a \right)} \right) + \left( {syn_{{g_j},{g_k}}^ + \left( {x_j^{p'},{g_k}\left( a \right)} \right) - syn_{{g_j},{g_k}}^ - \left( {x_j^{p'},{g_k}\left( a \right)} \right)} \right)} \right] \hfill \\
	\;\;\;\;\;\; \geqslant \left[ {{u_{{g_j}}}\left( {{g_j}\left( b \right)} \right) + {u_{{g_k}}}\left( {{g_k}\left( a \right)} \right) + \left( {syn_{{g_j},{g_k}}^ + \left( {{g_j}\left( b \right),{g_k}\left( a \right)} \right) - syn_{{g_j},{g_k}}^ - \left( {{g_j}\left( b \right),{g_k}\left( a \right)} \right)} \right)} \right] \hfill \\
	\;\;\;\;\;\; \geqslant \left[ {{u_{{g_j}}}\left( {{g_j}\left( b \right)} \right) + {u_{{g_k}}}\left( {{g_k}\left( b \right)} \right) + \left( {syn_{{g_j},{g_k}}^ + \left( {{g_j}\left( b \right),{g_k}\left( b \right)} \right) - syn_{{g_j},{g_k}}^ - \left( {{g_j}\left( b \right),{g_k}\left( b \right)} \right)} \right)} \right]. \hfill \\ 
	\end{gathered}
	\end{equation*}
	In case (c), utilizing the results from case (a), we can prove:
	\begin{equation*}
	\begin{gathered}
	\left[ {{u_{{g_j}}}\left( {{g_j}\left( a \right)} \right) + {u_{{g_k}}}\left( {{g_k}\left( a \right)} \right) + \left( {syn_{{g_j},{g_k}}^ + \left( {{g_j}\left( a \right),{g_k}\left( a \right)} \right) - syn_{{g_j},{g_k}}^ - \left( {{g_j}\left( a \right),{g_k}\left( a \right)} \right)} \right)} \right] \hfill \\
	\;\;\;\;\;\; \geqslant \left[ {{u_{{g_j}}}\left( {{g_j}\left( a \right)} \right) + {u_{{g_k}}}\left( {x_k^{q - 1}} \right) + \left( {syn_{{g_j},{g_k}}^ + \left( {{g_j}\left( a \right),x_k^{q - 1}} \right) - syn_{{g_j},{g_k}}^ - \left( {{g_j}\left( a \right),x_k^{q - 1}} \right)} \right)} \right] \hfill \\
	\;\;\;\;\;\; \geqslant \left[ {{u_{{g_j}}}\left( {{g_j}\left( a \right)} \right) + {u_{{g_k}}}\left( {x_k^{q'}} \right) + \left( {syn_{{g_j},{g_k}}^ + \left( {{g_j}\left( a \right),x_k^{q'}} \right) - syn_{{g_j},{g_k}}^ - \left( {{g_j}\left( a \right),x_k^{q'}} \right)} \right)} \right] \hfill \\
	\;\;\;\;\;\; \geqslant \left[ {{u_{{g_j}}}\left( {{g_j}\left( a \right)} \right) + {u_{{g_k}}}\left( {{g_k}\left( b \right)} \right) + \left( {syn_{{g_j},{g_k}}^ + \left( {{g_j}\left( a \right),{g_k}\left( b \right)} \right) - syn_{{g_j},{g_k}}^ - \left( {{g_j}\left( a \right),{g_k}\left( b \right)} \right)} \right)} \right] \hfill \\
	\;\;\;\;\;\; \geqslant \left[ {{u_{{g_j}}}\left( {{g_j}\left( b \right)} \right) + {u_{{g_k}}}\left( {{g_k}\left( b \right)} \right) + \left( {syn_{{g_j},{g_k}}^ + \left( {{g_j}\left( b \right),{g_k}\left( b \right)} \right) - syn_{{g_j},{g_k}}^ - \left( {{g_j}\left( b \right),{g_k}\left( b \right)} \right)} \right)} \right]. \hfill \\ 
	\end{gathered}
	\end{equation*}
	In case (d), utilizing the results from case (a), we can prove:
	\begin{equation*}
	\begin{gathered}
	\left[ {{u_{{g_j}}}\left( {{g_j}\left( a \right)} \right) + {u_{{g_k}}}\left( {{g_k}\left( a \right)} \right) + \left( {syn_{{g_j},{g_k}}^ + \left( {{g_j}\left( a \right),{g_k}\left( a \right)} \right) - syn_{{g_j},{g_k}}^ - \left( {{g_j}\left( a \right),{g_k}\left( a \right)} \right)} \right)} \right] \hfill \\
	\;\;\;\;\;\; \geqslant \left[ {{u_{{g_j}}}\left( {x_j^{p - 1}} \right) + {u_{{g_k}}}\left( {x_k^{q - 1}} \right) + \left( {syn_{{g_j},{g_k}}^ + \left( {x_j^{p - 1},x_k^{q - 1}} \right) - syn_{{g_j},{g_k}}^ - \left( {x_j^{p - 1},x_k^{q - 1}} \right)} \right)} \right] \hfill \\
	\;\;\;\;\;\; \geqslant \left[ {{u_{{g_j}}}\left( {x_j^{p'}} \right) + {u_{{g_k}}}\left( {x_k^{q'}} \right) + \left( {syn_{{g_j},{g_k}}^ + \left( {x_j^{p'},x_k^{q'}} \right) - syn_{{g_j},{g_k}}^ - \left( {x_j^{p'},x_k^{q'}} \right)} \right)} \right] \hfill \\
	\;\;\;\;\;\; \geqslant \left[ {{u_{{g_j}}}\left( {{g_j}\left( b \right)} \right) + {u_{{g_k}}}\left( {{g_k}\left( b \right)} \right) + \left( {syn_{{g_j},{g_k}}^ + \left( {{g_j}\left( b \right),{g_k}\left( b \right)} \right) - syn_{{g_j},{g_k}}^ - \left( {{g_j}\left( b \right),{g_k}\left( b \right)} \right)} \right)} \right]. \hfill \\ 
	\end{gathered}
	\end{equation*}
	To sum up, if $\Delta u_{{g_j}}^s + \sum\limits_{q = 1}^t {\left( {\eta _{{g_j},{g_k}}^{ + ,s,q} - \eta _{{g_j},{g_k}}^{ - ,s,q}} \right)}  \geqslant 0,\;\;s = 1,...,{\gamma _j},\;t = 1,...,{\gamma _k}$, $\forall \left\{ {{g_j},{g_k}} \right\} \in Syn$, then, for any pair of alternatives $a,b$ such that ${g_j}\left( a \right) \geqslant {g_j}\left( b \right)$ and ${g_k}\left( a \right) \geqslant {g_k}\left( b \right)$, $\forall \left\{ {{g_j},{g_k}} \right\} \in Syn$, we have
	\begin{equation*}
	\begin{gathered}
	\left[ {{u_{{g_j}}}\left( {{g_j}\left( a \right)} \right) + {u_{{g_k}}}\left( {{g_k}\left( a \right)} \right) + \left( {syn_{{g_j},{g_k}}^ + \left( {{g_j}\left( a \right),{g_k}\left( a \right)} \right) - syn_{{g_j},{g_k}}^ - \left( {{g_j}\left( a \right),{g_k}\left( a \right)} \right)} \right)} \right] \hfill \\
	\;\;\;\;\;\; \geqslant  \left[ {{u_{{g_j}}}\left( {{g_j}\left( b \right)} \right) + {u_{{g_k}}}\left( {{g_k}\left( b \right)} \right) + \left( {syn_{{g_j},{g_k}}^ + \left( {{g_j}\left( b \right),{g_k}\left( b \right)} \right) - syn_{{g_j},{g_k}}^ - \left( {{g_j}\left( b \right),{g_k}\left( b \right)} \right)} \right)} \right]. \hfill \\
	\end{gathered}
	\end{equation*}

	\begin{sidewaystable}[!htbp] \caption{\label{tab-app-1}Potential optimal values of $f\left( {v_{{g_j}}^p\left( a \right),v_{{g_j}}^p\left( b \right),v_{{g_k}}^q\left( a \right),v_{{g_k}}^q\left( b \right)} \right)$ and solutions at the optimum.}
		\centering
		\footnotesize
		\begin{tabular}{c c c c c c}
			\hline
			\multicolumn{1}{c}{No.} & \multicolumn{1}{c}{$v_{{g_j}}^p\left( a \right)$} & \multicolumn{1}{c}{$v_{{g_j}}^p\left( b \right)$} & \multicolumn{1}{c}{$v_{{g_k}}^p\left( a \right)$} & \multicolumn{1}{c}{$v_{{g_k}}^p\left( b \right)$} & \multicolumn{1}{c}{Value of objective function} \\
			\hline
			1     & $ - \frac{{\Delta u_{{g_k}}^q + \sum\limits_{s = 1}^{p - 1} {\left( {\eta _{{g_j},{g_k}}^{ + ,s,q} - \eta _{{g_j},{g_k}}^{ - ,s,q}} \right)} }}
			{{\eta _{{g_j},{g_k}}^{ + ,p,q} - \eta _{{g_j},{g_k}}^{ - ,p,q}}}$      & $ - \frac{{\Delta u_{{g_k}}^q + \sum\limits_{s = 1}^{p - 1} {\left( {\eta _{{g_j},{g_k}}^{ + ,s,q} - \eta _{{g_j},{g_k}}^{ - ,s,q}} \right)} }}
			{{\eta _{{g_j},{g_k}}^{ + ,p,q} - \eta _{{g_j},{g_k}}^{ - ,p,q}}}$      & $ - \frac{{\Delta u_{{g_j}}^p + \sum\limits_{t = 1}^{q - 1} {\left( {\eta _{{g_j},{g_k}}^{ + ,p,t} - \eta _{{g_j},{g_k}}^{ - ,p,t}} \right)} }}
			{{\eta _{{g_j},{g_k}}^{ + ,p,q} - \eta _{{g_j},{g_k}}^{ - ,p,q}}}$      & $ - \frac{{\Delta u_{{g_j}}^p + \sum\limits_{t = 1}^{q - 1} {\left( {\eta _{{g_j},{g_k}}^{ + ,p,t} - \eta _{{g_j},{g_k}}^{ - ,p,t}} \right)} }}
			{{\eta _{{g_j},{g_k}}^{ + ,p,q} - \eta _{{g_j},{g_k}}^{ - ,p,q}}}$      & 0  \\
			2     & 1     & 1     & 1     & 1     & 0 \\
			3     & 1     & 0     & 1     & 1     & $\Delta u_{{g_j}}^p + \sum\limits_{t = 1}^q {\left( {\eta _{{g_j},{g_k}}^{ + ,p,t} - \eta _{{g_j},{g_k}}^{ - ,p,t}} \right)}$ \\
			4     & 0     & 0     & 1     & 1     & 0 \\
			5     & 1     & 1     & 1     & 0     & $\Delta u_{{g_k}}^q + \sum\limits_{s = 1}^p {\left( {\eta _{{g_j},{g_k}}^{ + ,s,q} - \eta _{{g_j},{g_k}}^{ - ,s,q}} \right)}$ \\
			6     & 1     & 0     & 1     & 0     & $\Delta u_{{g_j}}^p + \sum\limits_{t = 1}^q {\left( {\eta _{{g_j},{g_k}}^{ + ,p,t} - \eta _{{g_j},{g_k}}^{ - ,p,t}} \right)}  + \Delta u_{{g_k}}^q + \sum\limits_{s = 1}^{p - 1} {\left( {\eta _{{g_j},{g_k}}^{ + ,s,q} - \eta _{{g_j},{g_k}}^{ - ,s,q}} \right)}$ \\
			7     & 0     & 0     & 1     & 0     & $\Delta u_{{g_k}}^q + \sum\limits_{s = 1}^{p - 1} {\left( {\eta _{{g_j},{g_k}}^{ + ,s,q} - \eta _{{g_j},{g_k}}^{ - ,s,q}} \right)}$ \\
			8     & 1     & 1     & 0     & 0     & 0 \\
			9     & 1     & 0     & 0     & 0     & $\Delta u_{{g_j}}^p + \sum\limits_{t = 1}^{q - 1} {\left( {\eta _{{g_j},{g_k}}^{ + ,p,t} - \eta _{{g_j},{g_k}}^{ - ,p,t}} \right)}$ \\
			10    & 0     & 0     & 0     & 0     & 0 \\
			\hline
		\end{tabular}
	\end{sidewaystable}
	\qed
\end{description}

\section*{Appendix C. Using the Choquet integral as a~preference model}
\noindent

\noindent In this section, we use the Choquet integral to represent interactions among criteria and illustrate how to adapt the proposed approach to such a preference model. Let ${2^G}$ denote the power set of $G$ (i.e., all subsets of criteria). A fuzzy measure (also called capacity) $\mu :{2^G} \to \left[ {0,1} \right]$ is defined such that it satisfies the following boundary and monotonicity conditions:
\begin{itemize}
	\item[(1a)] $\mu \left( \emptyset  \right) = 0$, $\mu \left( G \right) = 1$,
	\item[(1b)] $\mu \left( T' \right) \leqslant \mu \left( T \right)$ for all $T' \subseteq T \subseteq G$.
\end{itemize}
In this context, $\mu \left( T \right)$ for each $T \subseteq G$ can be interpreted as the importance weight assigned to the subset of criteria $T$. A useful representation of such a fuzzy measure is in terms of the M{\"o}bius transform:
\begin{equation*}
	\mu \left( T \right) = \sum\limits_{T' \subseteq T} {{m_\mu }\left( {T'} \right)},
\end{equation*}
for all $T \subseteq G$, where the M{\"o}bius representation ${{m_\mu }\left( {T'} \right)}$ is given by:
\begin{equation*}
	{m_\mu }\left( {T'} \right) = \sum\limits_{T \subseteq T'} {{{\left( { - 1} \right)}^{\left| {T' - T} \right|}}\mu \left( T \right)}.
\end{equation*}
A fuzzy measure is called $k$-additive if $\mu \left(  T  \right) = 0$ for every $T \subseteq G$ such that $\left| T \right| > k$. In practice, it is sufficient to consider 2-additive measures which means interactions for $n$-tuples ($n > 2$) are neglected. 

Given marginal function ${u_{{g_j}}}\left(  \cdot  \right)$ on each criterion $g_j$ and fuzzy measure $\mu \left(  \cdot  \right)$, for each alternative $a$, the Choquet integral is defined as:
\begin{equation*}
	{C_\mu }\left( a \right) = \sum\limits_{i = 1}^n {\left[ {{u_{{g_{\left( i \right)}}}}\left( {{g_{\left( i \right)}}\left( a \right)} \right) - {u_{{g_{\left( {i - 1} \right)}}}}\left( {{g_{\left( {i - 1} \right)}}\left( a \right)} \right)} \right]\mu \left( {{T_{\left( i \right)}}} \right)},
\end{equation*}
where $_{\left(  \cdot  \right)}$ is a permutation of the indices of criteria such that:
\begin{align*}
& {u_{{g_{\left( n \right)}}}}\left( {{g_{\left( n \right)}}\left( a \right)} \right) \geqslant {u_{{g_{\left( {n - 1} \right)}}}}\left( {{g_{\left( {n - 1} \right)}}\left( a \right)} \right) \geqslant  \cdots  \geqslant {u_{{g_{\left( 1 \right)}}}}\left( {{g_{\left( 1 \right)}}\left( a \right)} \right), \\
& {u_{{g_{\left( 0 \right)}}}}\left( {{g_{\left( 0 \right)}}\left( a \right)} \right) = 0, \;\;\;\;\text{and}\;\;\;\;{T_{\left( i \right)}} = \left\{ {{g_{\left( i \right)}},...,{g_{\left( n \right)}}} \right\}.
\end{align*}
The Choquet integral can also be defined in terms of the M{\"o}bius representation as:
\begin{equation*}
{C_\mu }\left( a \right) = \sum\limits_{T \subseteq G} {{m_\mu }\left( T \right)\mathop {\min }\limits_{{g_i} \in T} {u_{{g_i}}}\left( {{g_i}\left( a \right)} \right)}.
\end{equation*}

When using the Choquet integral as a preference model, the main problem is that performances on all criteria should be defined on the same scale or that the marginal values need to be given beforehand as the input of the problem. This is a strong assumption that limits the practical applicability of such a model. In this section, we adopt a scaling method to normalize all performances to the [0, 1] scale. A simple transformation to this end is given by:
\begin{equation*}
g{'_i}\left( a \right) = \frac{{{g_i}\left( a \right) - {\alpha _i}}}
{{{\beta _i} - {\alpha _i}}}.
\end{equation*}
The above transformation is problematic, because outliers would make the distribution skewed. Instead, we can use the following mapping:
\begin{equation*}
g{'_i}\left( a \right) = {F}\left( {{g_i}\left( a \right)} \right),
\end{equation*}
where $F$ is the cumulative distribution function such that $F: {g_i}\left( a \right) \rightarrow {\mathbf{P}}\left( {X \leqslant {g_i}\left( a \right)} \right)$. Function $F$ is not known and we can use the empirical distribution ${\hat F}$ of the training data to act as $F$ (i.e., $\hat F\left( {{g_i}\left( a \right)} \right)$ indicating the number of relative frequency of reference alternatives $a'$ such that ${g_i}\left( {a'} \right) \leqslant {g_i}\left( a \right)$). One needs to be aware that adopting such a scaling method to bring all performances to the common scale is again troublesome, because it assumes the marginal value functions on all criteria are limited to the same form, which is not appropriate for many problems. From the viewpoint of statistical learning, such a~scaling method would make the performance of learning model less advantageous, since it reduces the flexibility of an assumed preference model. However, since it is considered a standard way of performing such a translation, we will investigate its performance in the experimental study.
Using the above scaling method, the Choquet integral can be reformulated as follows:
\begin{equation*}
{C_\mu }\left( a \right) = \sum\limits_{T \subseteq G} {{m_\mu }\left( T \right)\mathop {\min }\limits_{{g_i} \in T} } g{'_i}\left( a \right).
\end{equation*}
With respect to a 2-additive fuzzy measure, we can rewrite ${C_\mu }\left( a \right)$ as:
\begin{equation*}
{C_\mu }\left( a \right) = {\mathbf{m}}_\mu ^{\text{T}}{{\mathbf{V}}_\mu }\left( a \right),
\end{equation*}
where ${{\mathbf{m}}_\mu }$ and ${{\mathbf{V}}_\mu }\left( a \right)$ are two vectors such that:
\begin{align*}
& {{\mathbf{m}}_\mu } = {\left( {{m_\mu }\left( {\left\{ {{g_1}} \right\}} \right),...,{m_\mu }\left( {\left\{ {{g_n}} \right\}} \right),{m_\mu }\left( {\left\{ {{g_1},{g_2}} \right\}} \right),...,{m_\mu }\left( {\left\{ {{g_{n - 1}},{g_n}} \right\}} \right)} \right)^{\text{T}}},\\
& {{\mathbf{V}}_\mu }\left( a \right) = {\left( {g{'_1}\left( a \right),...g{'_n}\left( a \right),\mathop {\min }\limits_{{g_i} \in \left\{ {{g_1},{g_2}} \right\}} g{'_i}\left( a \right),...,\mathop {\min }\limits_{{g_i} \in \left\{ {{g_{n - 1}},{g_n}} \right\}} g{'_i}\left( a \right)} \right)^{\text{T}}}.
\end{align*}
Then, we can adapt the proposed approach to the Choquet integral model and derive the following optimization problem:

\begin{align*}
({\text{P4}}): \;\;\;\;\;\; & Minimize\;  - d + {C_1}{\mathbf{m}}_\mu ^{\text{T}}{{\mathbf{S}}_\mu }{{\mathbf{m}}_\mu },\\ 
{\text{s}}{\text{.t}}{\text{.}} \;\;\;\;\;\; & {\mathbf{m}}_\mu ^{\text{T}}{{\mathbf{\Theta }}_{\mu ,k + 1}} \geqslant {\mathbf{m}}_\mu ^{\text{T}}{{\mathbf{\Theta }}_{\mu ,k}} + d,\;\;k = 1,...,q - 1,\\ 
& d \geqslant 0,\\
& \sum\limits_{{g_i} \in G} {{m_\mu }\left( {\left\{ {{g_i}} \right\}} \right)}  + \sum\limits_{{g_i},{g_j} \in G} {{m_\mu }\left( {\left\{ {{g_i},{g_j}} \right\}} \right)}  = 1,\\
& {m_\mu }\left( {\left\{ {{g_i}} \right\}} \right) \geqslant 0,\;\;{g_i} \in G,\\
& {m_\mu }\left( {\left\{ {{g_i}} \right\}} \right) + \sum\limits_{{g_j} \in T} {{m_\mu }\left( {\left\{ {{g_i},{g_j}} \right\}} \right)}  \geqslant 0,\;\;{g_i} \in G,\;T \subseteq G\backslash \left\{ {{g_i}} \right\},
\end{align*}
where ${{\mathbf{S}}_\mu } = \sum\limits_{k = 1}^q {\sum\limits_{a,b \in A_k^R} {\left( {{{\mathbf{V}}_\mu }\left( a \right) - {{\mathbf{V}}_\mu }\left( b \right)} \right){{\left( {{{\mathbf{V}}_\mu }\left( a \right) - {{\mathbf{V}}_\mu }\left( b \right)} \right)}^{\text{T}}}} }$ and ${{\mathbf{\Theta }}_{\mu ,k}} = \frac{1}
{{\left| {A_k^R} \right|}}\sum\limits_{a \in A_k^R} {{{\mathbf{V}}_\mu }\left( a \right)}$. The last three linear constraints ensure the boundary and monotonicity properties of the Choquet integral in case of using a~2-additive fuzzy measure.

\section*{Appendix D. Experimental results of the proposed approach in Section \ref{sec-experimental-analysis-2}}
\noindent
In the e-Appendix (supplementary material available on-line in form of a spreadsheet), we report the detailed results for the experimental study described in Section \ref{sec-experimental-analysis-2}. Specifically, we provide the average values of accuracy, precision, recall, and F-measure for different variants of the proposed approach applied to the nine accounted datasets. These variants are distinguished by the incorporated classification method, the way of modelling the interactions between criteria as well as an assumed number of linear pieces for all marginal value functions.
\end{document}